# ArchiveLens: A Generative Spatial Knowledge Framework for Semantic Reconstruction of Historical Maps

Yanan Wu[a], Yalin Yang[b], Chengbin Deng[c, d, *]
[a]*Department of Geography, The University of Central Arkansas, Conway, USA;* [b]*Department of Geology and Geography, West Virginia University, Morgantown, USA;* [c] Center for Spatial Analysis, *Norman, OK, USA;* [d] *Department of Geography and Environmental Sustainability, University of Oklahoma, Norman, OK, USA*
corresponding author: cdeng@ou.edu*

**Abstract**
Historical maps contain dense morphological and semantic information about past urban environments, but many digitization workflows focus primarily on extracting geometry, and do not capture cartographic semantics such as building use and construction materials. This study introduces ArchiveLens, a generative spatial knowledge framework that operationalizes historical maps as entity-centric, machine-queryable geospatial databases. The pipeline decouples morphological and semantic reconstruction through three stages: tiled YOLOv11-seg instance segmentation for building footprint extraction; deterministic material classification via hue, saturation, and value (HSV)-based color thresholds; and localized attribute parsing using a quantized vision–language model constrained by domain-specific prompts. Extracted entities are integrated into a hybrid Spatial–Temporal Retrieval-Augmented Generation (ST-RAG) architecture that combines dense semantic retrieval with symbolic spatio-temporal filtering and mandatory provenance grounding. Applied to 445 georeferenced Sanborn map sheets of Chicago (1918–1950), the framework achieves high geometric and semantic accuracy. Beyond performance, ArchiveLens advances GIScience by formalizing historical cartography as structured spatial knowledge representation, allowing researchers to query building attributes and compare spatial patterns across map sheets and years

## 1. Introduction

Historical maps constitute one of the most information-dense records of past landscapes (Harley, 1989). They record not only spatial geometry but also social organization, economic activity, risk assessment, and infrastructural systems. For urban and historical geographers, these documents are vital empirical instruments for tracing the morphological evolution of cities, analyzing the spatial distribution of industrialization, and uncovering the micro-geographies of neighborhood change. Through standardized cartographic symbol systems, color conventions, abbreviations, and typologies, historical cartography embeds structured semantic and classificatory systems within visual form (Annanias et al., 2023; O'Hara et al., 2024) . Accordingly, historical maps are not merely visual representations of spatial features but historically situated spatial knowledge systems (Harley, 1989).

As large-scale digitization efforts continue to convert historical cartographic archives into diverse digital resources, they have become increasingly accessible within contemporary spatial data infrastructures, significantly advancing accessibility, searchability, and interoperability (Justová & Cajthaml, 2023). These efforts have primarily addressed the geometric representation (e.g., segmentation, vectorization, and feature extraction) of mapped entities rather than the formalization of the structured semantic and classificatory systems embedded within cartographic conventions (Chen et al., 2024; Wu et al., 2023). Furthermore, historical archives are inherently fragmented, often published in localized, rolling volumes that resist continuous spatial or longitudinal analysis. This creates a persistent “semantic silence”: the richly encoded knowledge remains legible to human experts through close reading but is rarely instantiated as structured, machine-readable information at corpus scale, despite emerging end-to-end pipelines for extracting and linking map text/metadata from large collections (Grossner & Mostern, 2021; Kim et al., 2023). As a result, GIScience has limited ability to mobilize historical cartographic semantics for computational spatial analysis, reasoning, and modeling (Chen et al., 2024).

This study addressed this gap by introducing ArchiveLens, a three-stage generative spatial knowledge framework that translates the semantic and classificatory systems embedded in historical cartography into structured machine-readable geospatial records. First, we extracted discrete spatial features using high-precision instance segmentation. Second, we reconstructed multi-dimensional semantic and classificatory attributes using Vision-Language Models (VLMs). Third, we integrated these feature-level records and attributes into a provenance-grounded Spatio-Temporal Retrieval-Augmented Generation (ST-RAG) architecture, enabling semantically rich querying and reasoning over historical map corpora. We grounded and evaluated this framework using the Sanborn Fire Insurance Maps, a highly standardized and symbol-rich cartographic archive spanning 12,000 U.S. cities from the late nineteenth to mid-twentieth centuries. Regularized color schemes, abbreviation vocabularies, and dense building-level annotations make the Sanborn corpus a suitable empirical test case for modeling cartographic information. The methodological architecture of ArchiveLens is not Sanborn-specific: it offers a transferable computational paradigm for reconstructing the structured semantic systems embedded in other standardized historical cartographic collections.

This work advances historical GIS and GIScience in three main ways. First, it contributes to structured spatial knowledge representation and semantic GIS by demonstrating how historically embedded, practice-specific classificatory systems can be made explicit as interoperable, machine-readable spatial models at the feature level. Second, it extends historical GIS methodology from document-centric, manually curated interpretation toward automated, corpus-scale semantic enrichment, enabling new forms of longitudinal, multi-city, and multi-scale spatio-temporal analysis. Third, it proposes a knowledge-aware architecture for integrating generative AI into archival spatial infrastructures that emphasize epistemic transparency, provenance tracking, and reproducible feature-level grounding. These contributions help transform historical maps from passive background layers into active, formally modeled knowledge resources for GIScience.

## 2. Literature Review

### 2.1 The value of Historical Cartography and Sanborn Maps

Historical maps serve as indispensable spatial knowledge systems that capture high-resolution snapshots of past urban forms, socio-economic structures, and environmental conditions. In the context of the "Spatial Turn" across the humanities and social sciences, these documents have transcended their role as passive illustrations, emerging as primary evidence for reconstructing longitudinal urban dynamics(Withers, 2009). Scholars increasingly utilize these records as the foundation for "Deep Maps"—multi-layered spatial narratives that integrate physical geography with cultural and social history to create immersive understandings of past environments(Wei. Li, 2024). By georeferencing and digitizing these analog archives, researchers can trace patterns of urban growth and decline, analyze the impact of historical policy decisions, and preserve cultural heritage within modern Geographic Information Systems (Shen & Dong, 2022).

However, a significant research gap exists in the scalability and synthesis of historical maps. Current historical GIS (HGIS) scholarship frequently relies on manually georeferenced layers or localized case studies that do not generalize across heterogeneous archival corpora (Affek, 2013). While qualitative "deep mapping" offers rich narrative depth, it often lacks a standardized computational framework to manage large-scale spatial uncertainties or to synthesize cross-archive semantic data into a unified, machine-readable format (Haslhofer et al., 2013). Consequently, much of the latent spatial knowledge within these archives remains siloed within individual digital image files.

Among the vast corpus of historical cartography, the Sanborn Fire Insurance Maps represent one of the most comprehensive and standardized records of the American built environment from the late 19th to the mid-20th century. Originally produced to assist insurance underwriters in assessing fire risk, these maps offer an unparalleled level of building-level detail, including structural footprints, construction materials (e.g., frame, brick, stone), floor counts, and specific functional utilizations (Craine et al., 2023). Unlike traditional topographic maps, the Sanborn corpus provides a granular, entity-centric view of urban morphology, making it an essential resource for studying the localized impacts of urban renewal, highway construction, and industrial shifts (Sauder, 1980; Wrigley, 1949). Recent scholarship has leveraged these maps to create three-dimensional digital models of vanished neighborhoods, demonstrating their capacity to serve as a high-fidelity "time-memory" of urban space (Y. Lin et al., 2023).

Despite their detail, a second research gap arises in the formalization of these features as linked geographic entities. While existing automated methods have achieved high accuracy in identifying building footprints (geometric extraction), they frequently fail to formalize the structured semantic and cartographic information embedded in the source material (Sun et al., 2021). Existing workflows often treat building extraction and attribute classification as disjointed tasks, resulting in a loss of the "entity-centric" logic where a building's geometry is inextricably linked to its historical semantic profile (Uhl et al., 2020). Furthermore, there is a distinct lack of "grounded provenance" in current automated pipelines; information is often extracted into flat databases that sever the vital link back to the original cartographic evidence, making it difficult for historians to verify or audit the machine-generated data. Transforming these analog treasures into structured, entity-centric geospatial records remains a profound logistical and computational barrier.

## 2.2 Automated Extraction from Historical Maps

The automated processing of historical maps represents a long-standing challenge at the intersection of image analysis, GIS, and information retrieval (Chiang et al., 2019; Wei. Li, 2024). Early approaches relied on hand-crafted feature engineering (color thresholding, contour detection) to identify building outlines and text regions (Brown et al., 2012). Still, they were brittle to variations in scanning quality, symbolization, and map design. The advent of deep convolutional networks transformed the landscape, as Mask R-CNN demonstrated the feasibility of instance-level building segmentation from heterogeneous cartographic inputs (Garcia-Molsosa et al., 2021; Sakeena et al., 2023). Machine learning has also been applied to Sanborn map processing (Y. Lin et al., 2023), where support vector machines were used to extract building footprints and Mask R-CNN to identify building-level structures and materials, primary for 3D reconstruction rather than large-scale map parsing. Beyond Sanborn maps, deep learning methods have been widely applied to topographic map processing (Mattéo et al., 2021), cadastral map digitization (Timár & Biszak, 2010), and historical aerial photograph interpretation (Kindermann et al., 2023), confirming the maturity of pixel-wise and instance-wise segmentation pipelines while also revealing their limitations in terms of computational cost (Maxwell et al., 2020). In parallel, YOLO (You Only Look Once) has emerged as a leading single-stage detection framework that offers a favorable balance between detection accuracy and inference speed compared to two-stage detectors such as Faster/Mask R-CNN, making it attractive for large-scale geospatial workflows (Jiang et al., 2022). Several studies have shown that YOLO and its variants are well-suited to building-related tasks in remote sensing. For example, locally constrained YOLO (LOCO-YOLO) significantly improves the detection of small, densely distributed building footprints by tailoring the architecture and regression constraints to building size and spacing, outperforming related building detectors on large-area datasets (Xie et al., 2020). Other work demonstrates that customized YOLO models can match or surpass Mask R-CNN for separating individual buildings while being substantially faster at inference, which is critical for processing large map collections (Medjadi et al., 2024). These strengths motivate the choice of a YOLO-based architecture for extracting building footprints from historical maps in this study, in contrast to segmentation-only approaches that may achieve high pixel accuracy but lack the computational efficiency and object-centric representation needed for large-scale, information-rich historical GIS workflows.

## 2.3 Vision-Language Models for Semantic Information Understanding

Recent advances in Vision–Language Models (VLMs) have fundamentally reshaped document understanding by enabling joint reasoning over visual and textual signals. Models such as GPT-4V (Gallifant et al., 2024), LLaVA (B. Li et al., 2024), and Qwen-VL (Wang et al., 2024) are pretrained on large-scale image–text corpora and learn cross-modal representations that align visual regions with linguistic tokens. Unlike traditional pipelines that separate OCR, layout parsing, and semantic classification, these multimodal transformers perform image-conditioned text generation end-to-end, allowing structured information to be extracted directly from document images. Semantic meaning in historical maps is distributed across text, color encoding, symbol systems, and spatial adjacency. Consequently, interpretation requires the simultaneous integration of typographic transcription and spatial reasoning. Recent studies have demonstrated that domain-specific fine-tuning enhances the capacity of VLMs for map reading and cartographic understanding (Zhang et al., 2025). Additionally, structured prompt engineering has been shown to significantly improve reasoning reliability by constraining output schemas and guiding intermediate inference steps (Dang et al., 2022; Korzynski et al., 2023). Within historical GIS, VLMs therefore function not as autonomous interpreters but as controlled, domain-aware semantic reconstruction modules. Their methodological contribution lies in bridging geometric vectorization and feature-level semantic formalization, enabling the systematic translation of visually encoded cartographic attributes into machine-readable spatial knowledge structures.

### 2.4 Retrieval-Augmented Generation for Spatial Knowledge

Large language models (LLMs) pretrained on general-purpose corpora demonstrate strong linguistic and reasoning capabilities, but their knowledge of domain-specific geospatial concepts remains limited (Z. Lin et al., 2024). When confronted with complex geospatial modeling requirements, these models may generate outputs that appear plausible but are substantively incorrect, reflecting a lack of spatial grounding and domain alignment (Liang et al., 2025). In response to these limitations, Retrieval-Augmented Generation (RAG) can be conceptualized as a knowledge-grounded spatial reasoning framework in which generative models are explicitly anchored to retrieved geospatial evidence, thereby constraining inference within spatially referenced and externally verifiable information structures (Hu & Lu, 2025). In GIScience, RAG represents a framework for integrating generative language models with formally indexed spatial knowledge, thereby reconnecting probabilistic inference with geographically grounded information structures. R2RAG-Flood, a reasoning-reinforced training-free RAG framework for post-storm property damage nowcasting, conditions LLM predictions on geospatially retrieved reasoning trajectories, thereby demonstrating that RAG can substantially improve both accuracy and interpretability in spatially heterogeneous decision tasks (Huang et al., 2026). The same rationales extend to historical cartographic analysis. Historical map corpora are large, heterogeneous, and sheet-specific; without retrieval, a language model cannot reliably access the relevant local evidence needed for rigorous interpretation, nor can it provide a transparent evidentiary trail. RAG therefore supports historical-map research by (1) retrieving the most relevant sheets, regions, or extracted feature records for a given question, and (2) constraining synthesis to that retrieved evidence, enabling reproducible, provenance-aware answers. Building on these principles, our ArchiveLens AI agent adopts dense semantic retrieval (FAISS) to retrieve the most relevant archival records and then generates responses only from retrieved, citable evidence, effectively transforming the model from an open-ended generator into a grounded archival assistant.

## 3. Data Acquisition: The Old Insurance Maps Platform

The empirical foundation of this study is built upon historical Sanborn Fire Insurance Maps. While the original analog sheets are preserved and digitized by federal institutions such as the United States Library of Congress (LOC), these repositories typically distribute maps as isolated raster images. We utilized the Old Insurance Maps platform, a digital public humanities initiative that curates and hosts high-fidelity digital versions of these historical documents(Cox, 2023). By providing streamlined access to these archival assets, the platform facilitates efficient large-scale data acquisition and supports systematic corpus construction for computational analysis. Consequently, the map images obtained from this repository provide the visual clarity and geometric reliability necessary for training deep learning models.

## 4. Study Area

ArchiveLens is validated through a multi-year case study of Chicago, Illinois, encompassing key temporal snapshots from 1918, 1920, 1923, and 1950. Chicago represents one of the most spatially dense, systematically surveyed, and multi-year and extensively mapped urban environments available in the public archival domain, providing a robust empirical testbed for large-scale spatial-semantic reconstruction. This context ensures a sufficient volume of high-quality, spatially reliable reference data necessary for training, validation, and performance evaluation of the ArchiveLens framework. In total, 445 fully georeferenced Sanborn maps covering the City of Chicago form the empirical foundation of this study and constitute the spatial basis for subsequent geometric and semantic analysis. These Sanborn maps provide rich morphological, topological, and typographic information, enabling the execution of subsequent geometric vectorization and semantic extraction processes. Historically, Chicago underwent rapid and transformative urban development during the late nineteenth and early twentieth centuries, entering a period of accelerated industrialization, intensified land use, and the emergence of early high-rise commercial architecture within the central business district. This period of concentrated urban restructuring resulted in highly heterogeneous spatial configurations, including dense commercial cores, vertically stratified structures, mixed-use residential blocks, and expansive industrial corridors. The corresponding Sanborn maps encode this urban complexity through extreme cartographic density, layered symbolic conventions, and tightly packed structural geometries. This pronounced morphological and semantic heterogeneity provides an ideal stress-test environment for evaluating ArchiveLens, particularly its capacity to resolve overlapping geometries, delineate irregular building footprints, and disambiguate dense typographic annotations embedded within complex spatial layouts.

Figure 1 demonstrates the multi-scalar richness of information embedded in a typical 1918 Chicago Sanborn sheet. Figure 1(a) presents a neighborhood-scale view, capturing residential blocks, parcel boundaries, property lines, and the street network. Figure 1(b) illustrates parcel-level encoding, where each building is represented as a discrete polygon aligned with lot boundaries and annotated with standardized text attributes. Address numbers are positioned along street frontages, while internal labels encode function and structure, for example, “D” denotes a dwelling, and codes such as “1B” or “2B” indicate one- or two-story brick construction. These conventions effectively render the map an entity-centric database linking each footprint to its physical characteristics. Figure 1(c) emphasizes detailed spatial and topological relationships, integrating sequential address numbering, street names (e.g., “KAMERLING AV.”), and the contiguous arrangement of rowhouses, thereby documenting urban density and physical connectivity.

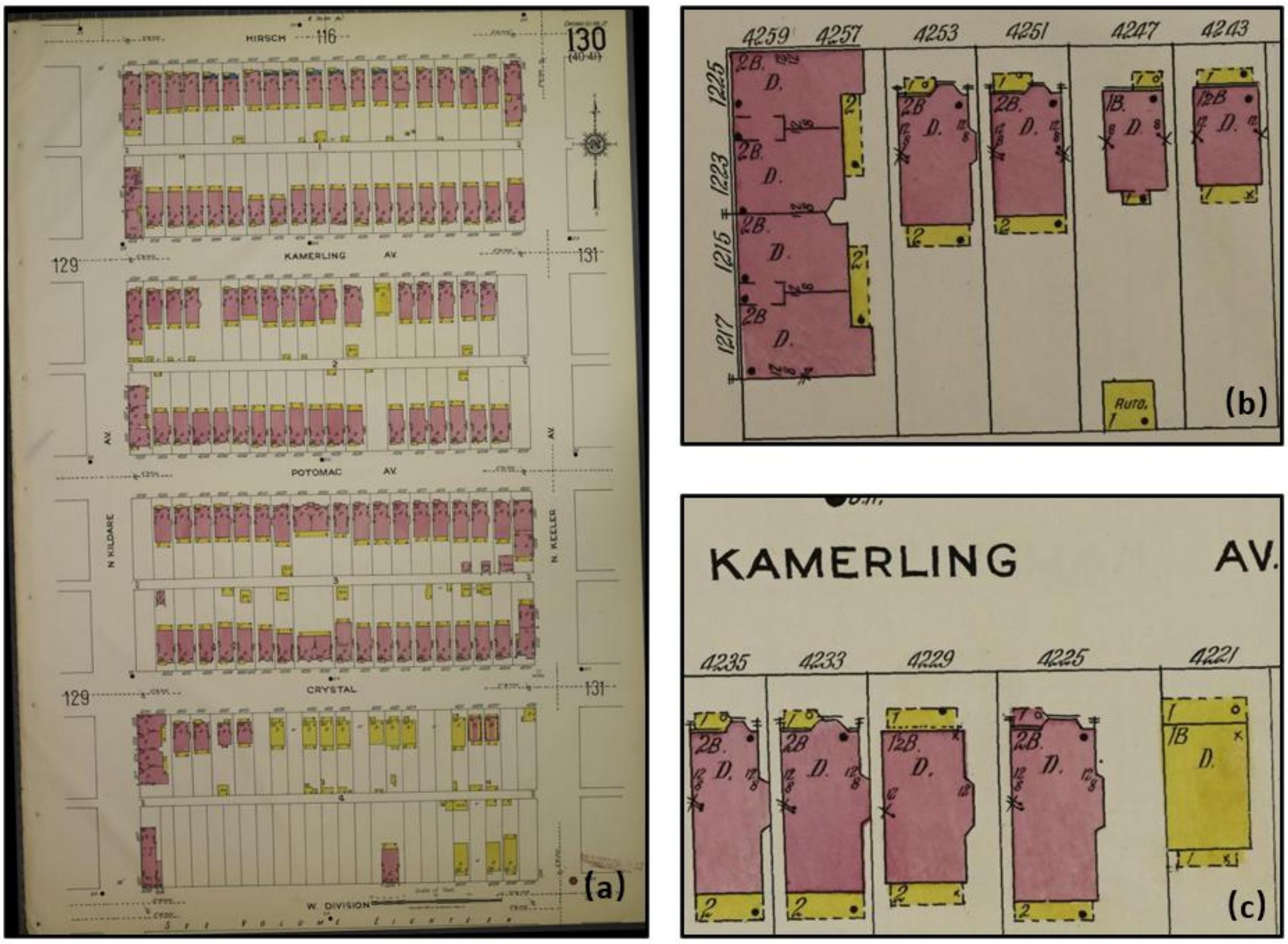

Figure 1. A 1918 Sanborn map for the City of Chicago in Illinois. (a) A full Sanborn sheet depicting multiple residential blocks, parcels/lot lines, street names, address ranges, and building footprints. (b) Parcel-level building representations and attribute encodings, including address numbers along the street frontage, discrete footprint polygons aligned to lot boundaries, and interior labels "D" for dwelling; "1B/2B" for one- or two-story brick construction. (c) A detailed view showing street naming ("KAMERLING AV."), sequential address numbering, construction/stories labels, and the spatial arrangement of adjacent rowhouse footprints across contiguous lots.

## 5. Methodology

ArchiveLens consists of three stages that extract urban morphological, reconstruct semantic information, and enable spatial reasoning (Figure 2)

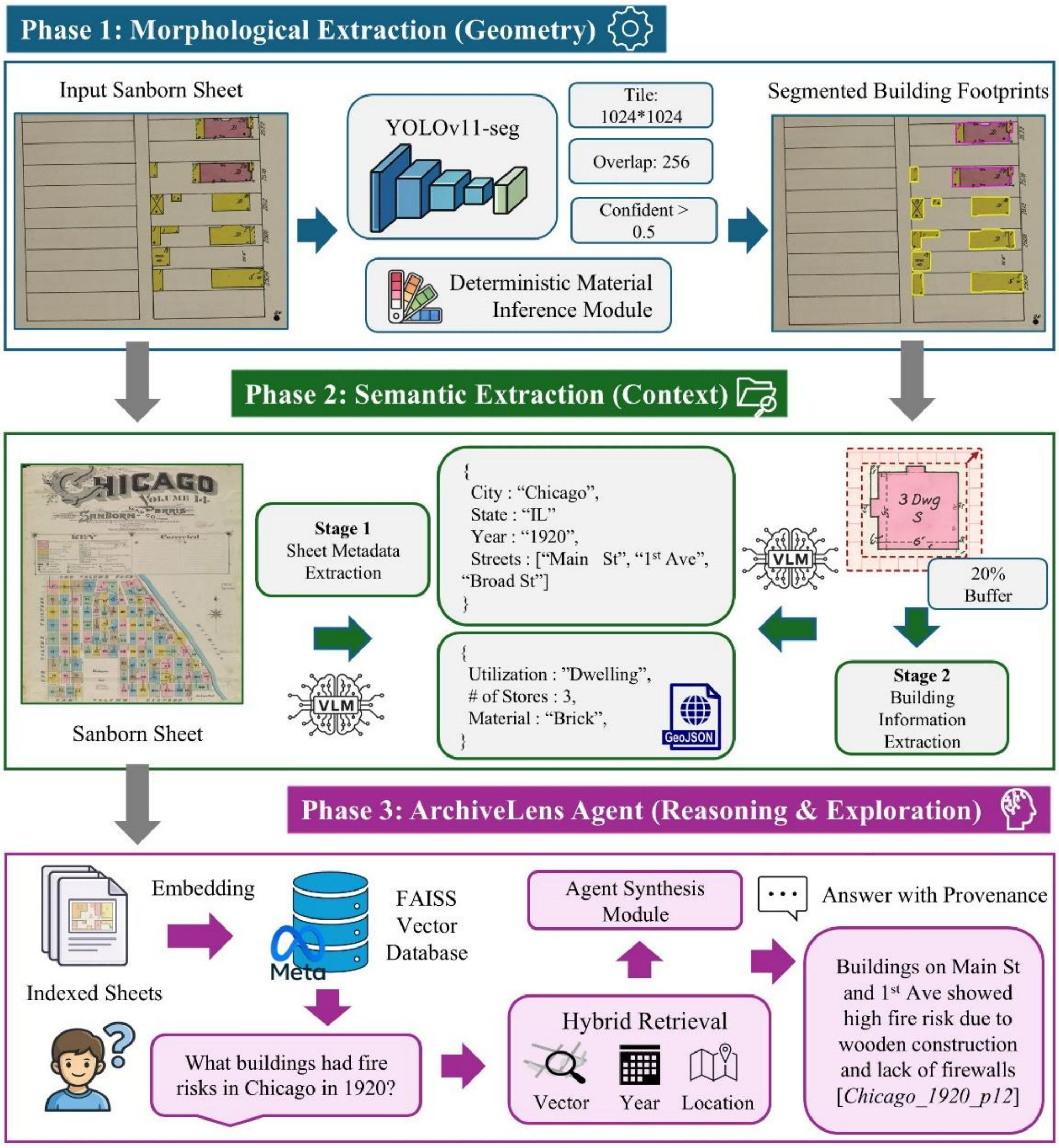

Figure 2. Architecture of ArchiveLens Framework. The pipeline comprises three phases: (1) Morphological Extraction, utilizing tiled YOLOv11-seg for precise building footprint delineation and spectral color inference for deterministic material classification; (2) Semantic Extraction, employing Vision-Language Models (VLMs) in a two-stage process to extract global sheet metadata and localized structure typography; and (3) The ArchiveLens Agent, functioning as a hybrid-retrieval spatial query engine that synthesizes vector embeddings, temporal filters, and spatial coordinates to answer complex historical geographic inquiries with clear provenance.

### 5.1 Geometric Context Extraction

The first phase of the ArchiveLens framework is to automate the vectorization of building footprints and the classification of their physical construction materials. This is executed through a localized instance segmentation model, followed by morphological refinement and deterministic spectral analysis.

#### 5.1.1 Building footprint extraction

Digitized Sanborn sheets at 300 DPI typically range from 5,000 x 6,000 to 10,000 x 12,000 pixels. Processing raster files of this magnitude natively exceeds the memory constraints of standard graphics processing units (GPUs) for direct instance segmentation. Consequently, we implemented a tiled inference strategy. The input rasters are systematically partitioned into 1024 x 1024 pixel overlapping windows. To ensure that structural footprints crossing tile boundaries are completely captured rather than arbitrarily truncated, a 256-pixel uniform stride overlap is enforced.

Because full-sheet inference relies on this tiled architecture, our model supervision and dataset generation must also operate at the tile level. We manually annotated building footprints across a representative sample of these historical map tiles to establish a reliable ground-truth dataset. The resulting corpus was partitioned using a standard 70:30 split, yielding 522 image tiles dedicated to model training and an independent holdout set of 240 tiles reserved for empirical testing. In addition, five complete Sanborn map sheets, including 420 tiles with 1024x1024 size, were reserved as an independent validation dataset. None of these validation sheets, or any tiles derived from them, appeared in either the training or testing datasets. Building footprint delineation is performed using a fine-tuned YOLOv11s-seg model, leveraging its decoupled head design and parallel upsampling branch for per-instance binary mask prediction. The model was initialized with COCO-pretrained weights and fine-tuned on the annotated Sanborn corpus using the configuration detailed in Table 1.

Table 1. YOLOv11s-seg Training Configuration

| Parameters | Values |
|---|---|
| Architecture | YOLOv11s-seg |
| Image/Tile Size | 1024 * 1024 pixels |
| Batch Size | 6 |
| Epochs | 300 |
| Optimizer | AdamW (lr = 0.01, weight decay = 0.0005) |
| Augmentation | Mosaic, color jitter, flip, rotation |
| Overlap | 256 pixels |

Segmentation is applied independently on each spatial tile. The overlap-induced inconsistencies between tiles are resolved by consolidating duplicate detections of the same building across adjacent partitions. Corresponding polygon masks are then unified to produce a single, consistent building footprint. The resulting geometries are transformed back into the original map coordinate system, thereby restoring a continuous and spatially coherent representation of the full sheet. We applied localized background filtering algorithms to eliminate false positive detections caused by dense cartographic line work (e.g., street annotations, parcel boundaries). First, the global sheet background color is estimated by randomly sampling 500 pixels exclusively from regions outside all predicted building footprints. For each predicted instance mask, the median BGR color is converted to the HSV color space. The normalized Euclidean distance between the mask's median HSV and the global background HSV is calculated. If $\geq 60\%$ of the pixels within a footprint match the background color profile within predefined tolerances ($\Delta H \leq 12, \Delta S \leq 40, \Delta V \leq 50$), the detection is classified as a cartographically induced false detection and removed. Following this validation stage, each mask is binarized at a 0.50 probability threshold. A morphological refinement pipeline is then executed to separate structurally distinct buildings that the segmenter erroneously merged. This consists of a morphological opening operation using an elliptical structuring element (kernel size = 3x3 pixels, 2 iterations), followed by a single-iteration erosion to smooth polygon vertices and align the boundaries with orthogonal architectural contours.

We evaluated the building-footprint detection performance of the fine-tuned models on a validation dataset of 420 tiles, treating the task as a single-class detection problem restricted to the "building" category. Model performance is quantified using established spatial evaluation metrics, including Precision, Recall, and mean Average Precision (mAP). We report mAP at an Intersection over Union (IoU) threshold of 0.50 (mAP50), as well as the more stringent averaged metric computed across IoU thresholds ranging from 0.50 to 0.95 (mAP50–95). These metrics are calculated independently for bounding boxes and polygon masks, enabling simultaneous assessment of geometric localization accuracy and pixel-level boundary accuracy.

5.1.2 Deterministic Material Inference

Sanborn maps encode structural building materials through a standardized cartographic color ontology. This attribute data is deterministically extracted via a Region of Interest (ROI) spectral classification applied to each validated footprint. As illustrated in Figure 3, the workflow translates analog cartographic symbology into defined volumetric thresholds within a 3D HSV topological model. For a given building, the bounding box of the instance mask is extracted from the global raster. To avoid spectral contamination from surrounding line work or neighboring structures, the analysis is strictly constrained to pixels where the predicted binary mask equals 1. The BGR values of these strictly in-mask pixels are converted to the HSV color space, and their median Hue, Saturation, and Value are computed. The dominant material class is then assigned by evaluating these median coordinates against the deterministic lookup boundaries defined in Table 2.

Table 2. HSV Topological Thresholds for Material Classification

| Material | Color | H Range (°) | S Range (%) | V Range (%) |
|---|---|---|---|---|
| Frame (Wood) | Yellow | 20–45 | 40–100 | 40–100 |
| Brick | Pink/Red | 340–360, 0–20 | 25–100 | 35–100 |
| Stone | Blue | 190–250 | 20–100 | 30–100 |
| Adobe | Gray | Any | 0–20 | 25–75 |

The five complete Sanborn sheets served as an independent validation dataset and were paired with manual XML annotations of building materials to rigorously quantify the performance of the spectral inference module. During the validation phase, computationally predicted footprints are spatially matched to these human-annotated ground-truth polygons using an Intersection over Union (IoU) threshold of 0.30. Classification performance is subsequently evaluated using two distinct metrics: building-level accuracy (the raw proportion of correctly classified structures) and an area-weighted accuracy. This area-weighted approach approximates pixel-level Region of Interest (ROI) fidelity by scaling the correctness of each prediction by its bounding box footprint, ensuring that massive industrial structures and minor outbuildings are evaluated proportionally to their spatial dominance.

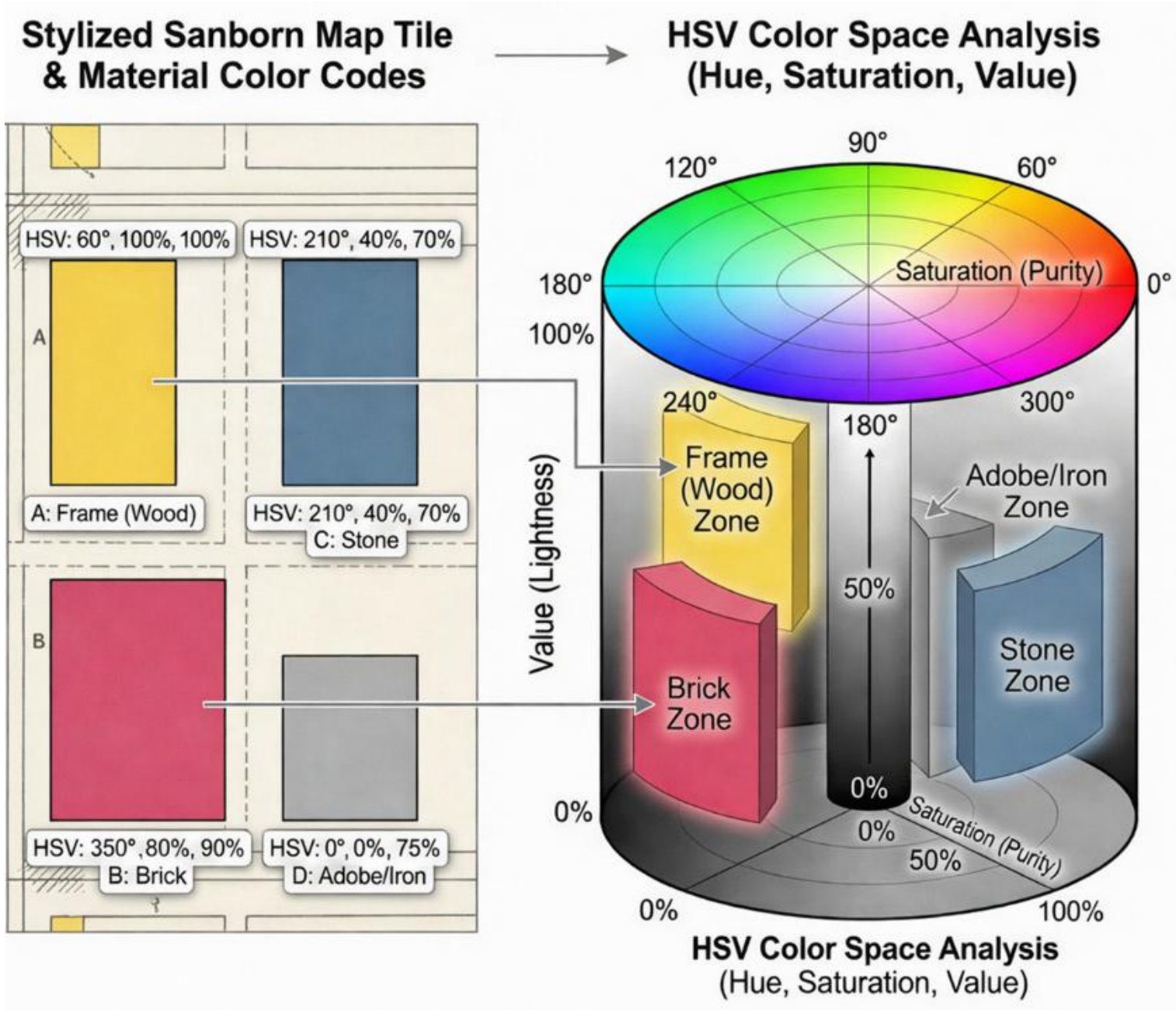

Figure 3. HSV-based classification of building construction materials. Standardized analog cartographic symbology in the source map (left) is transformed into predefined volumetric thresholds within a 3D HSV topological model (right).

5.2 Semantic Attribute Extraction

The second stage bridges the gap between geometry and semantics by extracting typographic information from map imagery and associating it with the building footprints generated in Stage 1. Semantic extraction was performed using Qwen3-VL-8B-Instruct (Bai et al., 2025), loaded with 4-bit NormalFloat (NF4) quantization, double quantization, FP16 computation, automatic device mapping, and scaled dot-product attention. Full-sheet images were resized to 1,280 pixels and generated with a maximum of 1,024 new tokens, whereas building crops were resized to 512 pixels and generated with a maximum of 512 new tokens. Sampling was disabled for both stages (do_sample=False); therefore, temperature and top-p sampling were not applied. Each sheet and building crop was processed once in the reported corpus extraction.

5.2.1 Global Sheet-Level Metadata Extraction

The first stage establishes the macro-geographic context by extracting sheet-level metadata, including the city, state, publication year, volume, and sheet number. The full-resolution map sheet is downsampled using a high-quality anti-aliasing filter (Lanczos resampling) to accommodate the VLM's optimal spatial token limits without sacrificing the legibility of fine typographic details (such as title cartouches and street names). A significant challenge with VLMs in historical document analysis is the propensity for spatial and temporal hallucinations. We introduce a "Filename Prior Prompting" strategy to overcome the hallucinations. Archival repositories systematically encode ground-truth geographic metadata into the digital filenames of the maps (e.g., identifying the city, state, and year). The extraction pipeline dynamically injects this filename string into the VLM's system prompt. By explicitly instructing the model to use the textual filename as a high-confidence anchor while visually validating the map imagery, we significantly constrain the model's generation space, forcing it to output a highly accurate, predefined JSON schema of the map's geography (Appendix A.1).

### 5.2.2 Building-Level Attribute Extraction

The second stage extracts building-level attributes, such as business names and the number of stories, for the individual geometries extracted in Phase 1. Processing the entire map sheet with the VLM for building-specific information introduces substantial visual noise and increases the risk of semantic cross-contamination among adjacent structures. To mitigate these issues, we employ a localized extraction strategy.

For each vectorized building, a targeted raster crop is generated. To ensure the crop captures the internal structural text while safely excluding the typography of neighboring buildings, we compute a dynamic contextual buffer margin based on the dimensions of the predicted bounding box:

$$b_x = clip\left(w_{bbox} * r_{buf}, b_{min}, b_{max}\right) \quad (1)$$

$$b_y = clip(h_{bbox} * r_{buf}, b_{min}, b_{max}) \quad (2)$$

Where $w_{bbox}$ and $h_{bbox}$ represent the bounding box width and height, $r_{buf} = 0.20$ is the dynamic buffer ratio, and the absolute pixel margins are clamped between a minimum ($b_{min}$) and maximum ($b_{max}$) threshold.

This adaptive mathematical scaling is critical for cartographic processing because it applies proportionally larger contextual borders to small outbuildings, where text often spills over the geometric footprint, while maintaining tight constraints on massive industrial structures to avoid encapsulating adjacent parcels. The dynamically cropped image tensors are passed to the VLM alongside a zero-shot prompt. To prevent free-text hallucinations and align the model with historical cartographic standards, the prompt explicitly restricts the model's output distribution using a controlled vocabulary of standard Sanborn abbreviations, detailed in Table 3.

Table 3. Standardized Sanborn Building Utilization Abbreviations

| Abbreviation | Semantic Meaning | Cartographic Context |
|---|---|---|
| D or Dwg | Dwelling | Standard residential structures |
| F | Flat | Multi-family residential units or apartments |
| S | Store | General commercial or retail spaces |
| A or Auto | Auto / Garage | Automotive storage or repair facilities |
| Po | Post Office | Post Office facilities |
| Sal | Saloon | Bars, taverns, or public houses |
| B | Basement | Subterranean structural levels |
| W | Warehouse | Industrial or commercial storage |
| Mfg | Manufacturing | Industrial production facilities |

Because VLMs can occasionally generate malformed structural outputs, a fault-tolerant parsing algorithm is applied to the generated text. This ensures that valid semantic key-value pairs are successfully recovered even if the model produces minor typographic variations or unclosed brackets. Finally, these parsed semantic attributes are definitively linked to the Phase 1 geometric records using their shared spatial coordinates, creating a fully enriched, vectorized dataset.

Similar to the validation protocol established for material inference, the accuracy of the VLM-based utilization extraction is evaluated against a manually annotated ground-truth dataset. Because historical buildings may accommodate multiple uses, such as a commercial storefront with residential units above, building-utilization extraction was treated as a multilabel classification task. Given the imbalanced class distribution, performance was assessed using exact-match accuracy、micro-F1、macro-F1、precision、recall、confusion matrix.

### 5.3 ST-RAG Architecture (Knowledge Retrieval and Synthesis)

The final phase of the framework operationalizes the extracted geometric and semantic data into a queryable, Spatial-Temporal Retrieval-Augmented Generation (ST-RAG) system. This agent synthesizes dense semantic retrieval with deterministic spatio-temporal filtering to answer complex urban history inquiries while maintaining strict archival provenance.

#### 5.3.1 Spatio-Temporal Knowledge Indexing

The heterogeneous geometric and semantic outputs from Phase 1 and Phase 2 are serialized into flat, sheet-level textual documents to construct the retrieval corpus. Because dense embedding models inherently prioritize the dominant semantic tokens in a document, we engineered a specific serialization hierarchy. Explicit spatio-temporal anchors (city, state, and year) are forcefully prepended to the beginning of every sheet's textual representation (e.g., *"This is a Sanborn Fire Insurance Map of Chicago, Illinois from the year 1950"*), followed by the aggregated list of building utilizations, street names, and extracted typographic notes. Each serialized document $D_i$ is mapped to a high-dimensional dense vector space using a pre-trained Sentence Transformer model, $f_\theta$:

$$E_i = f_\theta(D_i) \in R^d \quad (3)$$

These embeddings are indexed using a Facebook AI Similarity Search (FAISS) architecture. To optimize for computational efficiency during retrieval, the vectors undergo $L^2$ normalization, allowing the system to compute the cosine similarity between a query vector ($E_q$) and document vectors ($E_i$) using a highly optimized inner product search:

$$sim(q, D_i) = \frac{E_q \cdot E_i}{\|E_q\| \|E_i\|} \quad (4)$$

Records were embedded using sentence-transformers/all-MiniLM-L6-v2. Embeddings were converted to 32-bit floating-point vectors, L2-normalized, and indexed using FAISS inner-product search. Because the vectors were normalized before indexing, inner product is equivalent to cosine similarity for ranking.

#### 5.3.2 Intent-Aware Hybrid Retrieval

When a user submits a natural language query, pure semantic retrieval is often insufficient; a query asking for "industrial buildings in Chicago in 1950" requires exact temporal and spatial matching, which dense vectors alone approximate poorly. ArchiveLens employs an intent-aware hybrid retrieval mechanism to overcome this limitation. The intent parser is rule-based and does not use a learned intent-classification model. Regular expressions and controlled dictionaries extract explicit four-digit years, recognized city names, volume and sheet identifiers, quoted entity phrases, and requested-attribute intents. These parsed constraints determine which indexed records are eligible for retrieval. Upon receiving a query, an intent-parsing module executes heuristic evaluations to extract explicit constraints: a temporal set $T_{query}$ (four-digit years), a spatial set $S_{query}$ (target cities matching a predefined gazetteer), and a Boolean semantic intent flag $I_{query}$ triggered by comparative or structural keywords (e.g., "compare," "factory," "landmark").

Retrieval proceeds in two coordinated stages. First, an initial broad candidate pool is constructed using FAISS-based Approximate Nearest Neighbor (ANN) retrieval over sentence-level embeddings. This stage prioritizes semantic recall rather than precision, ensuring that conceptually relevant map sheets are not prematurely excluded due to lexical mismatch. By expanding the candidate set (typically $N \geq 200$), the system mitigates the well-known brittleness of pure vector similarity in historically noisy textual corpora. Second, the semantically recalled candidate set is subsequently refined through Boolean masking derived from parsed query constraints (e.g., year, city, or intent-specific requirements). This stage enforces exact geographic and temporal alignment while preserving semantic ranking within the filtered subset, thereby integrating symbolic logic with vector-based similarity in a principled hybrid retrieval framework.

Mathematically, the final retrieved set $R$ is defined as the subset of the broad dense recall pool $P$ that satisfies the parsed heuristic constraints:

$$R = \{D_i \in P \mid (D_{year} \in T_{query}) \wedge (D_{city} \in S_{query}) \wedge (D_{has_buildings} \in I_{query})\} \quad (5)$$

If the query parser detects multiple years or multiple cities, indicating a comparative historical analysis, the system dynamically scales the baseline $K$ threshold to ensure a balanced temporal and spatial distribution in the final context window.

### 5.3.3 Provenance-Grounded Generative Synthesis

The final phase uses the locally hosted Qwen3-VL-8B-Instruct model to synthesize the retrieved historical data into a coherent natural language response. The most critical constraint in archival AI systems is the mitigation of hallucination; historians require an unbroken chain of evidence. To enforce this, the synthesis LLM is governed by a strict "Mandatory Citation" system prompt (see Appendix A.3 for details). The prompt requires LLM to act strictly as a data synthesizer over the retrieved context block. Notably, the system prompt mandates that any factual claim, building reference, or structural description generated by the model must be immediately followed by its source sheet identifier in brackets (e.g., [chicago_ill_1950_vol_11_p28]). This design choice transforms the LLM from a probabilistic text generator into a grounded archival assistant, ensuring that every piece of spatial and historical information provided to the user can be deterministically traced back to the original cartographic TIFF file and its corresponding vectorized geometries.

## 5.4 Evaluation Design

### 5.4.1 Agent Retrieval Benchmark

We evaluated the spatial capabilities of ArchiveLens Agent using a controlled benchmark that separates record retrieval from answer generation. The evaluation tests whether the agent can identify a target building record, retrieve the associated Sanborn evidence, recover an explicitly indexed attribute or sheet-level context, cite the supporting sheet, and refrain from unsupported inference when the requested attribute is not represented in the Sanborn sheet.

We randomly sampled 500 buildings using seed 20260724 and generated no more than three questions per building, producing 1,500 questions. The benchmark contains 1,200 answerable questions and 300 unavailable-attribute questions. The answerable questions cover building utilization (n=401), source provenance (n=408), construction material (n=165), sheet-level road context (n=167), record details (n=48), storey count (n=9), and entity presence (n=2). The 300 unavailable questions request owner, architect, builder, or original construction-date information (n=75) for each category. Following standard top-$k$ retrieval evaluation in open-domain question answering (Karpukhin et al., 2020) , we measured the proportion of questions for which the designated gold target appeared among the first $k$ retrieved records, with $k$ in $\{1,3,5\}$. Because the archive has a hierarchical sheet–building structure, retrieval was evaluated at two levels of granularity, analogous to sheet- and building-level evaluation in hierarchical retrieval (Liu et al., 2021; Zhao et al., 2024).

#### 5.4.2 Retrieval and Response Evaluation Metrics

We assessed how ArchiveLens interprets and uses the retrieved evidence along four complementary dimensions: spatial reasoning, topological awareness, semantic reconstruction, and hallucination mitigation. Spatial reasoning refers to the ability to interpret and combine directional, metric, locational, and temporal relationships among geographic entities (Stogiannidis et al., 2025). Pyo et al. (2026) evaluate cartographic reasoning using topological, metric, and directional relations and require multi-step interpretation of map legends, scales, labels, orientation, and geometries. MapEval (Dihan et al., 2025) similarly evaluates compositional geospatial reasoning through text-based questions. We referred to these benchmarks when defining the evaluation dimensions for ArchiveLens and adapted their concepts to georeferenced, building-level historical archives. Topological awareness refers to the ability to recognize qualitative spatial relationships that remain meaningful without requiring exact measurements, such as whether two objects are disconnected, touching, overlapping, intersecting, or whether one object contains or lies within another. Ji et al. (2025) evaluates foundation models using formal topological predicates, including equal, disconnected, externally connected, partially overlapping, and tangential or non-tangential containment relations. ArchiveLens adopts this interpretation to examine relationships among building footprints. Semantic reconstruction measures whether ArchiveLens can translate historical cartographic conventions, such as colors, abbreviations, and labels into normalized, machine-readable building attributes. Hallucination mitigation measures whether every factual statement in the response is supported by retrieved map evidence and linked to the correct source sheet. It also evaluates whether the agent appropriately indicates uncertainty or declines to answer when the requested information is absent, unreadable, or outside the scope of the map.

#### 5.4.3 Cross-Year Query Evaluation

A cross-year comparison task was designed to assess the temporal query capability of ArchiveLens using the query: “Compare the extracted residential, commercial, and industrial records between 1918 and 1920.”. The purpose of this evaluation was to test whether the agent could correctly identify multiple publication years, apply year-specific temporal filters, construct separate record subsets, normalize extracted building-utilization codes into broader analytical categories, and summarize the resulting distributions in a structured comparison.

For each requested year, ArchiveLens independently retrieved and aggregated the corresponding indexed building records. Extracted Sanborn utilization codes were normalized into three broad analytical categories: residential, commercial, and industrial. Records that did not contain utilization codes belonging to these categories, including basements and other functional types, were retained in an “outside requested categories” group rather than excluded from the analysis. The evaluation also incorporated an explicit spatial-coverage constraint. Because the Sanborn sheets available for different publication years do not necessarily represent identical geographic extents, the resulting year-specific summaries were treated as cross-sectional comparisons of the indexed records rather than direct measures of longitudinal change within the same locations.

5.4.4 Retrieval Strategy Comparison

We conducted a controlled comparison to examine whether the hybrid ST-RAG retrieval strategy offers advantages over semantic or symbolic retrieval used independently. Three retrieval configurations were compared: dense-only retrieval, keyword/symbolic filter-only retrieval, and the full ST-RAG approach. The comparison focused on how effectively each strategy retrieved supporting evidence for queries involving functional descriptions, recorded building attributes, and temporal or map-identification constraints.

Dense-only retrieval relies exclusively on semantic similarity between the user query and the indexed records. It does not explicitly enforce metadata constraints such as city, year, volume, or sheet number. Keyword/symbolic filter-only retrieval instead uses explicit metadata and keyword constraints extracted from the query, prioritizing records that match information such as city, year, volume, sheet number, and relevant query terms without using semantic similarity. ST-RAG combines these two strategies by first identifying semantically relevant candidate records and then applying metadata constraints and phrase matching to refine the results. This design allows the system to consider both conceptual similarity and the specific geographic and temporal requirements expressed in the query.

The three retrieval configurations were evaluated using a 50-question set. The questions covered public facilities, industrial functions, storage and commercial uses, map-specific localization queries and semantic extraction. Spatial scope was represented through indexed city, volume, and sheet metadata. The benchmark included questions from the 1918, 1920, 1923, and 1950 editions, represented by 19, 9, 10, and 12 questions, respectively. Among the 50 questions. Retrieval performance was evaluated at the question level using the top-5 reference-candidate retrieval rate. A question was counted as successfully retrieved as long as the response in top-5 matched a preidentified supporting record. The overall retrieval rate was calculated as the proportion of the 50 questions meeting this criterion. Metadata compliance was evaluated separately at the record level by examining whether each returned record satisfied the city, year, volume, and sheet constraints identified from the corresponding query.

## 6. Results

### 6.1 Performance of Geometric Context Extraction

The efficacy of the morphological extraction phase was quantitatively evaluated using an independent validation dataset comprising 420 spatial tiles. For bounding box localization, the fine-tuned YOLOv11s-seg architecture achieved a Precision of 0.9264 and a Recall of 0.9018 (Figure 4). The mean Average Precision at an IoU threshold of 0.50 ($mAP_{50}$) reached 0.9361, indicating that the pipeline successfully identified and localized the vast majority of historical structures. When evaluated across the more stringent IoU threshold gradient of 0.50 to 0.95 ($mAP_{50\text{-}95}$), the bounding box performance maintained a score of 0.8147. Beyond bounding box localization, the model's capacity to delineate precise polygonal boundaries (instance segmentation) yielded commensurate metrics. The pixel-level mask Precision was recorded at 0.9302, alongside a mask Recall of 0.9051. The segmentation $mAP_{50}$ closely mirrored the bounding box performance at 0.9328, while the strict segmentation $mAP_{50\text{-}95}$ settled at 0.7816.

Following geometric delineation, the accuracy of the deterministic spectral classification module was evaluated. Within the validation dataset, 890 manually annotated ground-truth buildings contained explicit construction material labels. To assess attribute accuracy independently of minor boundary disagreements, predicted footprints were spatially joined to these ground-truth polygons using a baseline IoU threshold of 0.30, resulting in 782 matched building instances. Operating exclusively within these matched geometries, the deterministic HSV color space analysis achieved a building-level classification accuracy of 0.905 (708 correct material predictions out of 782). When accounting for the relative footprint size of different structures via an area-weighted metric (approximating pixel-level Region of Interest fidelity), the classification accuracy increased to 0.924. This positive performance delta suggests that the spectral inference is particularly reliable on massive, prominent structures (such as industrial warehouses) where the cartographic color fill is contiguous and well-preserved. Misclassifications were disproportionately concentrated among smaller outbuildings, where typographic overlaps or scanning artifacts disrupted the localized color space.

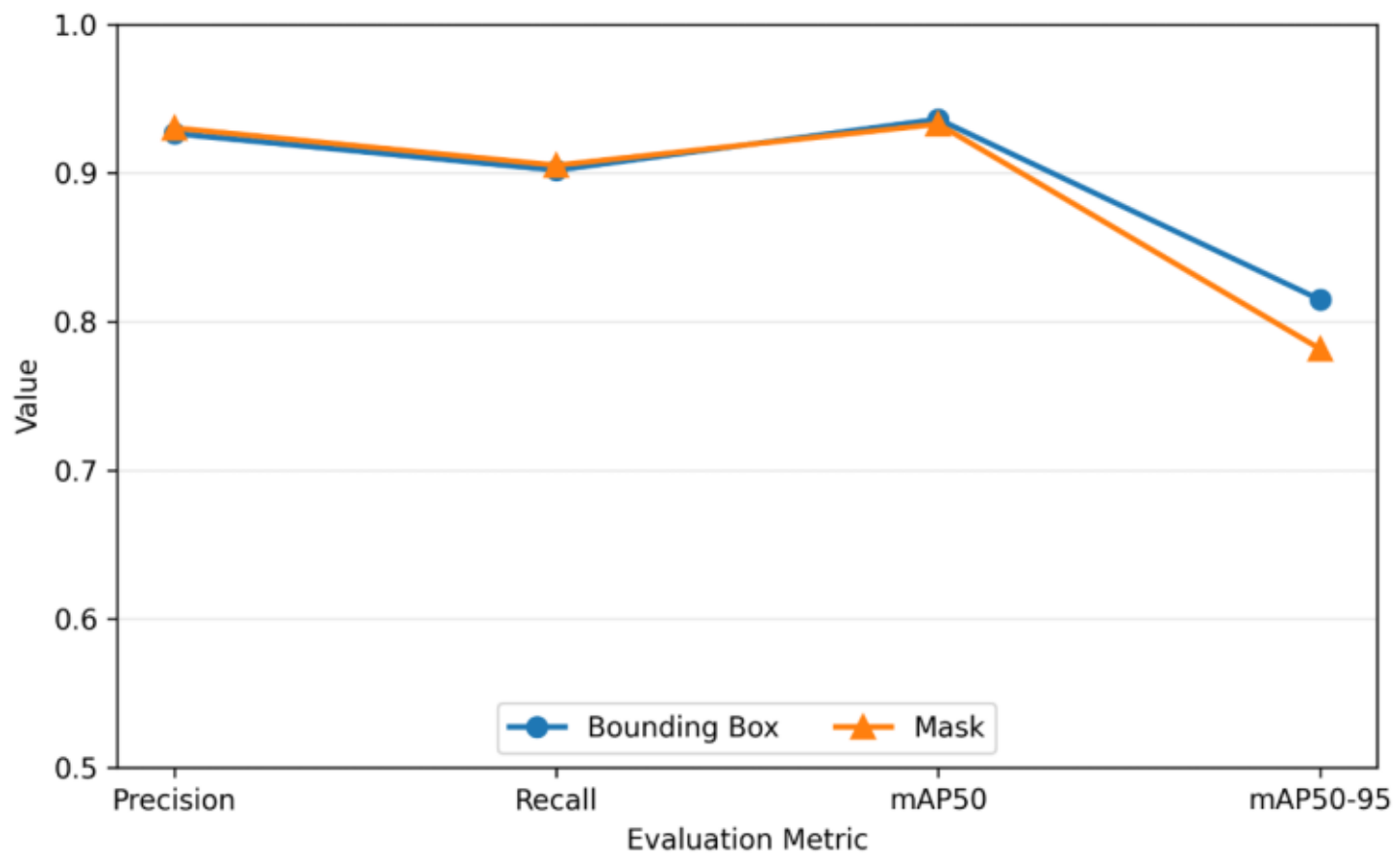

Figure 4. Performance comparison for building detection and instance segmentation. Precision, recall, mAP50, and mAP50–95 are reported separately for bounding-box localization and polygon-mask segmentation.

As illustrated in Figure 5, two notable capabilities emerged during inference. First, despite the manual ground-truth dataset being strictly constrained to orthogonal bounding boxes (left panel), the network successfully extrapolated pixel-level morphological contours. Rather than replicating rectangular constraints, the model predicted precise polygonal masks that conform to actual architectural footprints (right panel). Second, in several large industrial complexes, the predicted masks merged adjacent annotated units into larger continuous footprints. This behavior illustrates a difference between the annotation unit and the model output and should be interpreted with respect to the intended definition of a building entity.

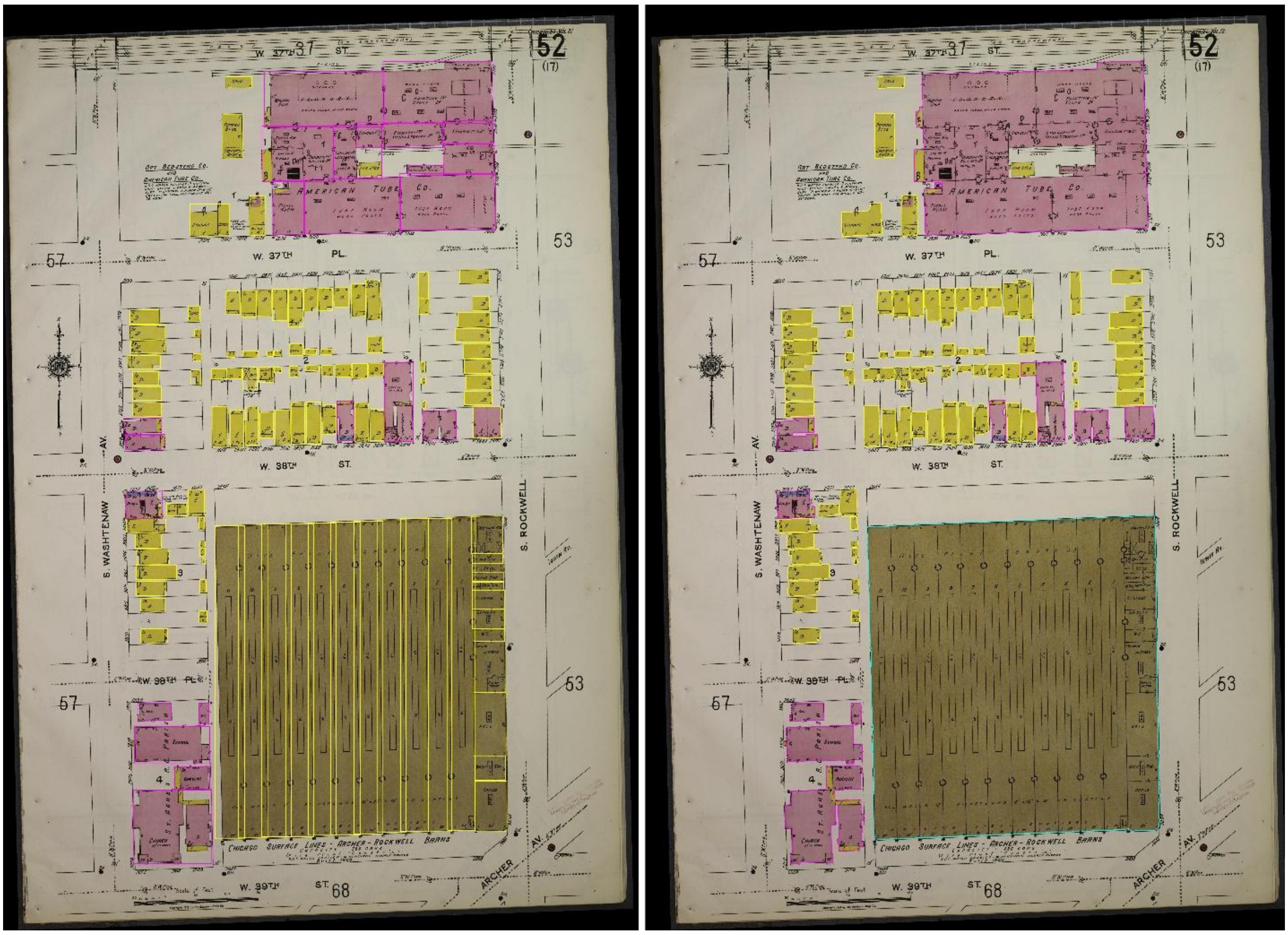

Figure 5. A visualization of an industrial map sheet. The left panel displays the manual ground-truth annotations utilizing fragmented, class-agnostic bounding boxes. The right panel displays the model's inference, demonstrating pixel-level polygonal segmentation, the geometric unification of massive industrial structures, and the deterministic spectral assignment of construction material.

The scalability of the ArchiveLens framework was demonstrated by applying the pipeline across a broader corpus of unannotated Chicago map sheets spanning four temporal intervals: 1918, 1920, 1923, and 1950. The resulting structural extraction are detailed in Table 4.

Table 4. Morphological Extraction Yields: Building Construction Materials by Publication Year

| Material | 1918 | 1920 | 1923 | 1950 | Total |
|---|---|---|---|---|---|
| Brick (Pink) | 2070 | 829 | 4087 | 4271 | 11257 |
| Frame (Yellow) | 5018 | 5770 | 7386 | 6562 | 24736 |
| Adobe (Gray) | 6 | 92 | 514 | 109 | 721 |
| Stone (Blue) | 27 | 65 | 134 | 930 | 1156 |

6.2 Performance of Semantic Attribute Extraction

The second phase of the evaluation quantifies the framework's ability to accurately extract typographic information using the quantized Vision-Language Model (Qwen3-VL-8B-Instruct). Performance was assessed across both the macro-geographic (sheet-level) and micro-geographic (building-level) extraction stages.

The efficacy of the first stage of semantic extraction was evaluated via a manual audit of twenty randomly map sheets. The VLM, constrained by the "Filename Prior Prompting" strategy, achieved an extraction accuracy of 100% for the foundational macro-geographic parameters, including city name, state, publication year, volume, and sheet number. The second stage evaluates the VLM's capacity to parse localized typographic codes (e.g., dwelling, store, flat) from dynamically generated bounding box crops. The evaluation utilized an independent dataset comprising 627 manually annotated building utilization labels.

Because historical structures may serve mixed-use functions (e.g., a commercial storefront on the ground floor with residential flats above), utilization extraction was evaluated as a multi-label classification task. Considering the multi-label classification and imbalance data, we employed exact-match accuracy, micro- and macro-F1 scores, per-class precision and recall, and confusion patterns for less frequent classes to measure performance. The utilization classification achieved an exact-match accuracy of 85.33% ($\frac{535}{627}$), with a micro-F1 of 0.853 ($\frac{1070}{1254}$) and a macro-F1 of 0.739. Figure 6 displayed the manual annotation and VLM prediction. Class-specific recall varied substantially (Figure 7), with Dwelling achieving 93.0%, Unknowm achieving 89.5%, and Auto 78.3%. Post Office had the lowest recall (40.0%), followed by Saloon (57.1%), Store (60.9%), Flat (62.5%), Warehouse (66.7%), and Manufacturing (72.2%). Errors in these less frequent classes predominantly involved predictions of Dwelling or Unknown. This pattern suggests difficulty distinguishing minority utilization classes, although the confusion matrix alone cannot establish whether these errors arose from typographic ambiguity or other factors. Estimates for classes with few reference instances, particularly Post Office (n = 5), should be interpreted cautiously.

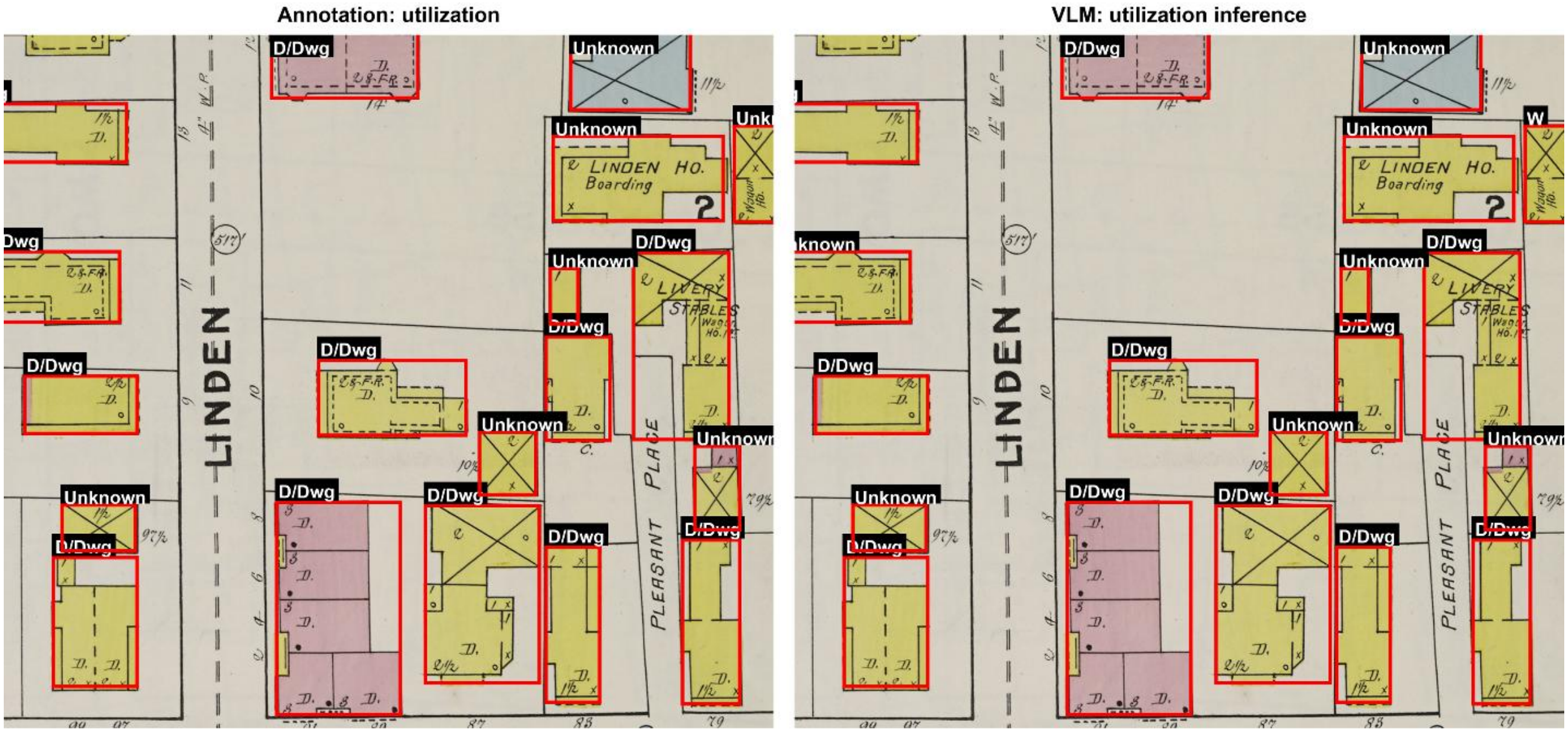

Figure 6. A comparative visualization of semantic utilization extraction. The left panel shows manual ground-truth utilization labels enclosed in red bounding boxes, while the right panel displays the corresponding zero-shot VLM predictions (e.g., 'D/Dwg' for dwelling, 'W' for warehouse).

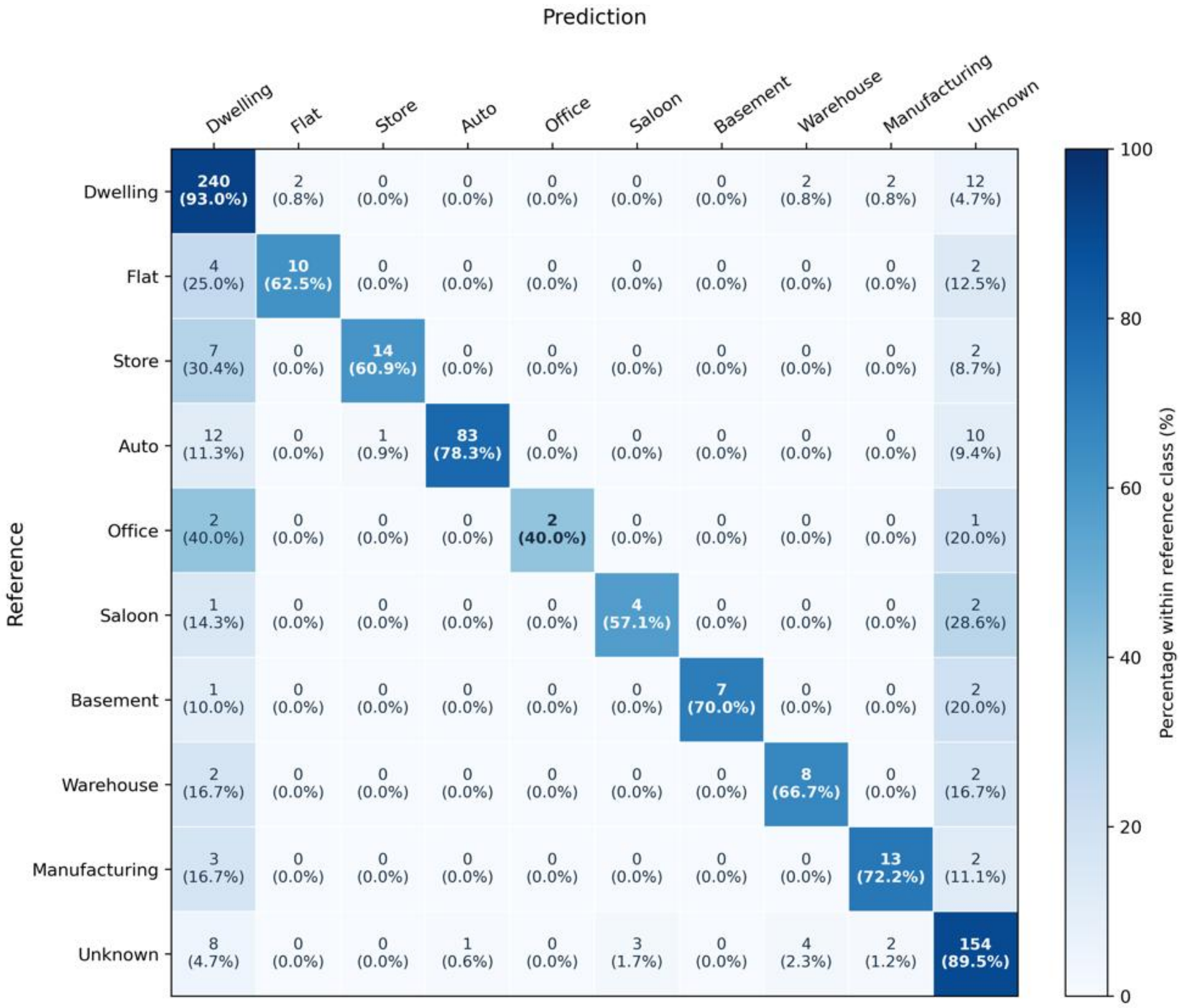

Figure 7. Confusion matrix for building-utilization identification. Rows represent reference labels and columns represent model predictions. Each cell reports the count and percentage within its reference class, with diagonal percentages indicating per-class recall. Color intensity represents row-normalized percentages. Unknown indicates an unassigned utilization label.

Figure 8 illustrates several limitations of VLM-based building-utilization identification. In panel (a), the model assigned D/Dwg to a building bearing an "A" annotation, failing to distinguish between dwelling and auto. In panel (b), a building explicitly labelled "S" was similarly classified as D/Dwg. Comparable errors were observed in other obliquely oriented buildings, suggesting that building orientation and associated text rotation may affect recognition, although their effects were not evaluated independently. Panel (c) shows a traction substation incorrectly classified as S (store), revealing both a mismatch between the recorded function and the predicted code and a limitation of the predefined vocabulary, which lacks a dedicated category for electrical infrastructure. In panel (d), the prediction "D, S, B" recovered dwelling and store labels but omitted the visible "Sal." annotation and combined basement information with functional uses. Such incomplete predictions were also observed in buildings containing multiple utilization codes within dense, complex textual layouts. Together, these cases highlight challenges in abbreviation interpretation, recognition of explicitly named functions, and complete multilabel extraction, underscoring the importance of accounting for text orientation and density while distinguishing structural attributes from functional categories during extraction and evaluation.

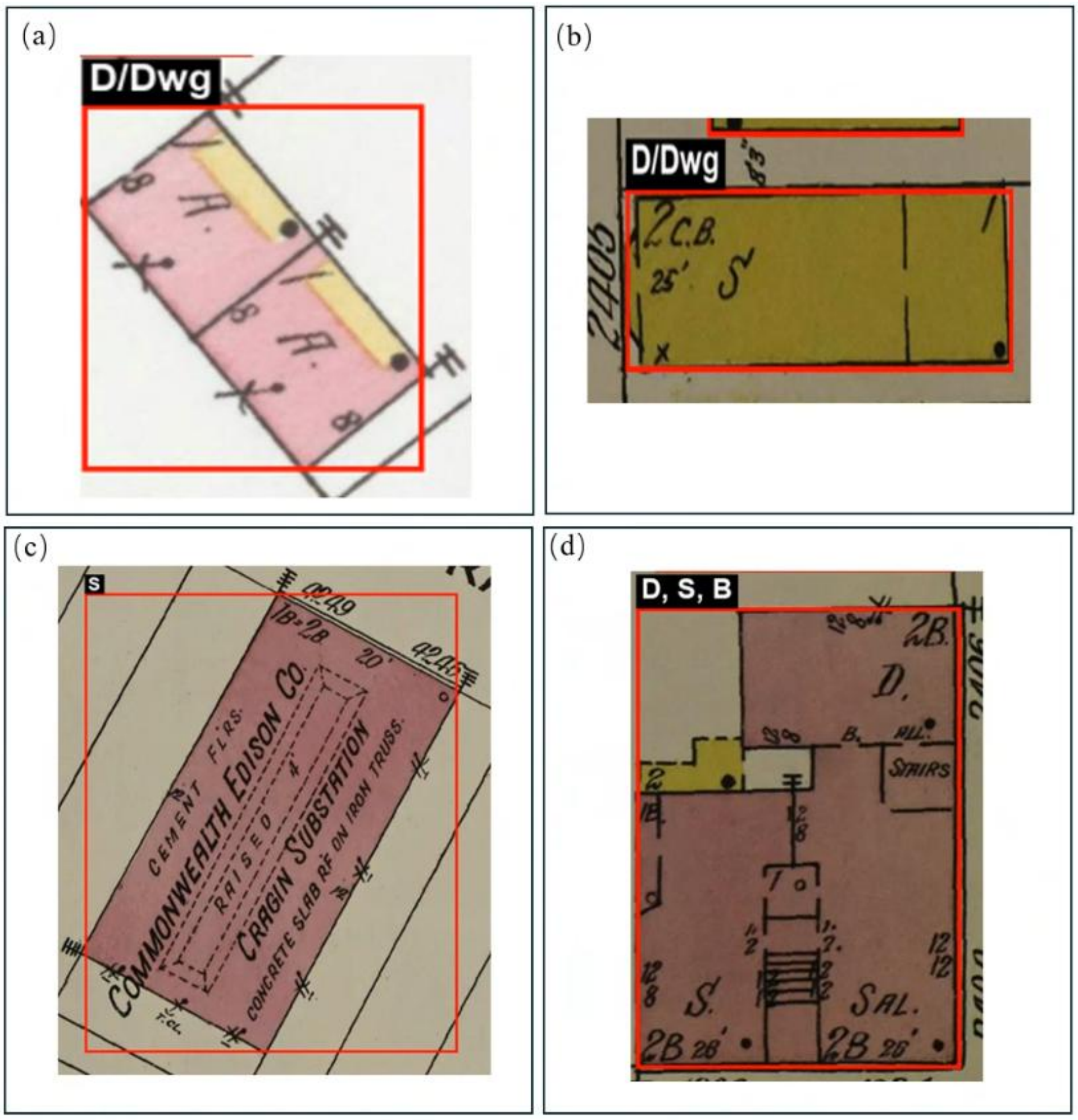

Figure 8. Examples of VLM errors in building-utilization identification**.** (a) An auto-annotated "A" classified as D/Dwg; (b) a building labelled "S" classified as D/Dwg; (c) a traction substation classified as S (store); and (d) incomplete multilabel extraction, with "D, S, B" omitting the visible "Sal." annotation and including basement information. Red boxes indicate predicted building bounding boxes, and black labels display predicted utilization codes.

Across the processed sheets, the framework successfully extracted functional utilization codes for 37,870 buildings. The number of utilization codes is more than the number of buildings, as a single building may have multiple utilization codes.

Table 5. Semantic Extraction Yields: Building Utilization Codes by Publication Year

| Utilization | 1918 | 1920 | 1923 | 1950 | Total |
|---|---|---|---|---|---|
| Dwelling | 5,582 | 3,207 | 6,571 | 8,052 | 23,412 |
| Flat | 116 | 14 | 157 | 233 | 520 |
| Store | 369 | 282 | 225 | 143 | 1,019 |
| Auto | 489 | 250 | 1,152 | 3,932 | 5,823 |
| Office | 12 | 6 | 9 | 7 | 34 |
| Saloon | 50 | 2 | 3 | 4 | 59 |
| Warehouse | 89 | 61 | 36 | 23 | 209 |
| Manufacturing | 71 | 31 | 22 | 27 | 151 |
| Basement | 32 | 14 | 15 | 32 | 93 |
| Unknown | 1981 | 3,143 | 4,040 | 2,523 | 11,687 |

These tables summarize extraction outputs rather than historical trends. Sanborn sheets from different publication years cover different parts of Chicago, so variations in building counts and attributes reflect differences in spatial coverage as well as building density. The results demonstrate the pipeline's scalability; longitudinal interpretation requires comparisons within a consistent geographic area.

6.3 ST-RAG Architecture

The output generated by ArchiveLens Agent demonstrates the integration of the hybrid Spatial-Temporal Retrieval-Augmented Generation (ST-RAG) pipeline. As demonstrated in Figure 9, the agent successfully retrieved a highly relevant 1918 map sheet detailing the "Federal Steel Fixture Co." in response to a user query asking to "Introduce the buildings related to industry in Chicago". The generated text effectively synthesizes macro-level geographic data with micro-level structural attributes (e.g., a "Manufacturing" utilization code and specific internal features such as "Sheet metal works," "wood posts & berms," and a "cement floor").

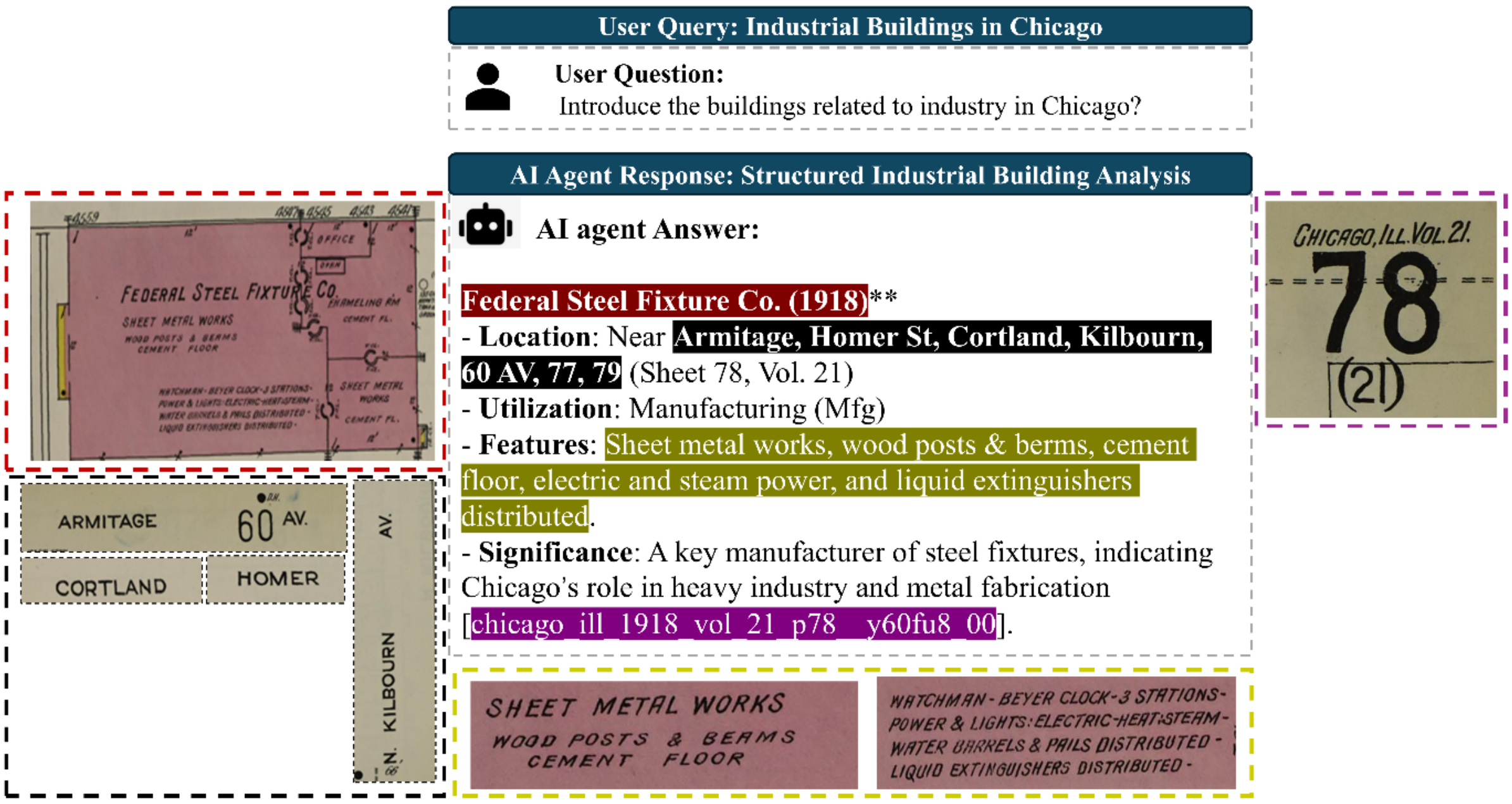

Figure 9. Qualitative evaluation of the ArchiveLens Agent's provenance-grounded synthesis and topological awareness. The system processes a query regarding industrial buildings in Chicago by extracting data from a 1918 map sheet. The response contextualizes the building within its local street network and appends a mandatory digital citation linking directly to the source cartography.

Notably, this synthesis extends beyond isolated semantic retrieval by demonstrating an operational understanding of urban topology. As previously noted, drawing meaningful historical conclusions from Sanborn archives requires bounding data within standardized spatial constraints. The agent directly resolves this requirement. As shown in Figure 9, the model did not merely extract the isolated footprint of the factory; it accurately contextualized the facility within its bounding street network, explicitly delineating its location relative to Armitage, Homer St, Cortland, and Kilbourn. This topological awareness is achieved because the underlying vector index explicitly binds the localized building geometries to the global sheet-level street metadata extracted during Phase 2. By organizing the data topologically rather than merely lexically, the agent enables researchers to execute spatially constrained queries, such as isolating industrial structures strictly along a specific street corridor across multiple decades, thereby overcoming the fragmented, rolling-volume publication schedule of the original physical maps.

Furthermore, this result validates the successful enforcement of the "Mandatory Citation" system prompt. The synthesis module grounds its output by concluding the factual summary with a precise source sheet identifier ([chicago_ill_1918_vol_21_p78__y60fu8_00]). The citation mechanism provides source traceability and enables verification of generated claims, but citation presence alone does not guarantee factual correctness.

### 6.4 Evaluation Results

#### 6.4.1 Agent Retrieval Benchmark Results

Based on the controlled benchmark described in Section 5.4.1, ArchiveLens demonstrated strong retrieval performance at both the sheet and building levels. Gold-sheet recall counts a query as successful when at least one of the top-$k$ records belongs to the designated gold sheet, whereas exact-building recall requires the specific gold building record to occur within the top $k$. Across all 1,500 questions, gold-sheet recall was 96.8% at rank 1, 98.8% at rank 3, and 99.1% at rank 5. Exact-building recall was 95.0%, 97.3%, and 98.1%, respectively.

#### 6.4.2 Retrieval and Response Evaluation Results

For each response, we manually traced the generated claims back to the original map source and validated the results. The example in Figure 10 demonstrates how the ArchiveLens agent integrates the four evaluated capabilities within a single response. Spatial reasoning is reflected in the identification of the downtown area near South Michigan Avenue and South Wabash Avenue as a concentration of commercial activity, combining geographic location, street names, and the 1950 map year. Topological awareness appears in the way the response situates the mapped buildings within the street-bounded area centered on Michigan and Wabash Avenues. Semantic reconstruction is shown through the interpretation of historical map labels such as “Auto Stage,” “Pit Garage,” and “Beer Depot” as evidence of automotive, storage, and commercial functions, which are then synthesized into a description of a commercial and mixed-use district. Hallucination mitigation is supported by the inclusion of the publication year, volume, sheet number, and source identifier, allowing the answer to be traced to the original Sanborn sheet. Based on map evidence showing a local concentration of commercial and mixed-use buildings, ArchiveLens inferred that this area represented Chicago’s “historic downtown core” and the “epicenter of retail, transportation, and office space.” This interpretation constitutes a broader contextual inference rather than a conclusion directly established by the cited map sheet.

The response also reveals two minor errors in interpreting Sanborn abbreviations. The label “Auto St’ge” on the map denotes “Automobile Storage,” but ArchiveLens transcribed it as “Auto Stage.” Similarly, “Priv. Garage,” meaning “Private Garage,” was incorrectly read as “Pit Garage.” These errors show that the model can recognize that abbreviated labels encode building functions and can use them for semantic interpretation, but it does not always reconstruct the abbreviations accurately. This limitation is particularly important for Sanborn maps, where shortened terms, stylized handwriting, faded printing, and closely spaced letters are common. The result therefore demonstrates partial semantic-reconstruction capability while also indicating the need for additional training or domain-specific abbreviation dictionaries to improve the model's recognition and normalization of Sanborn terminology.

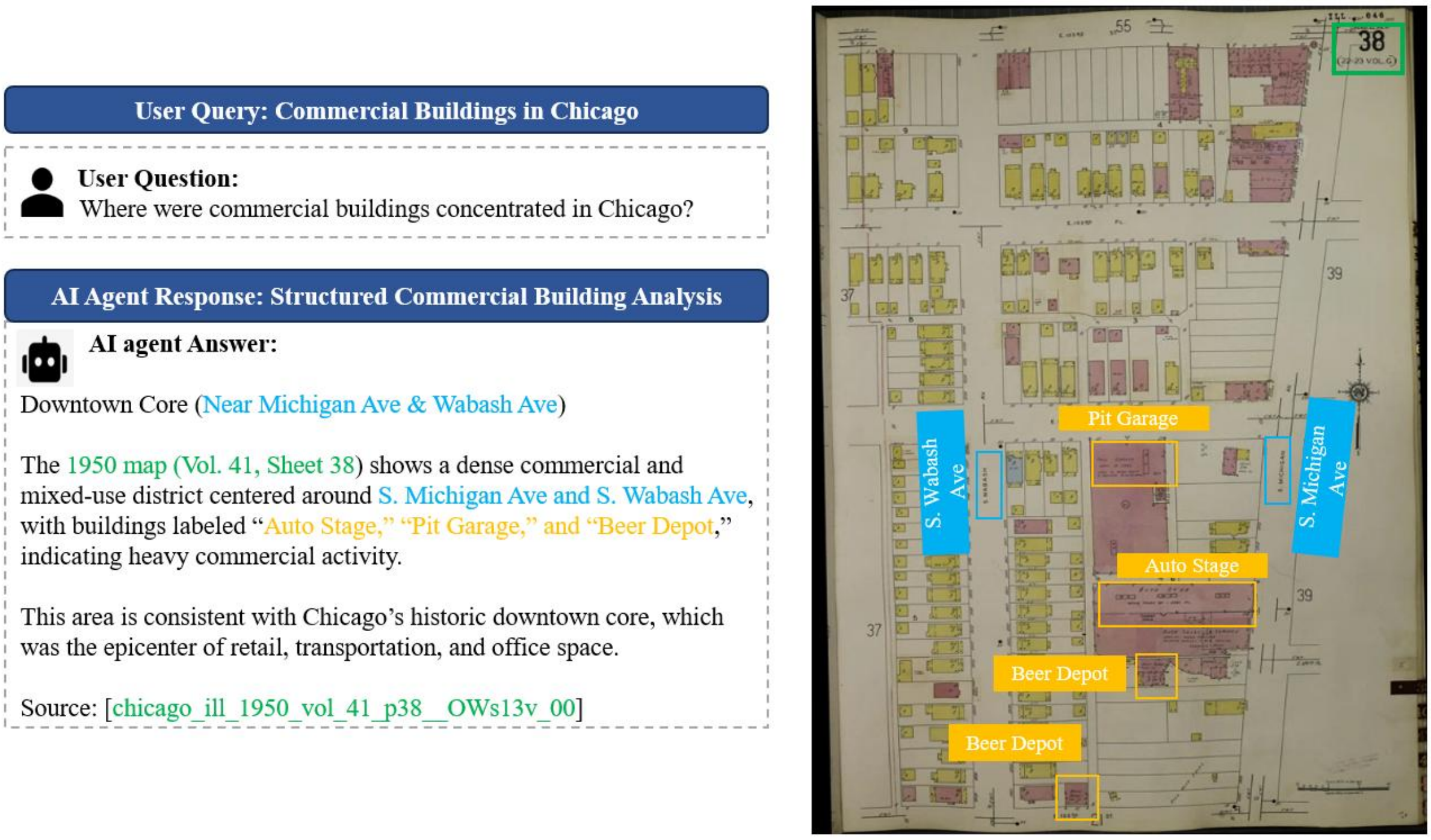

Figure 10. A response from ArchiveLens on the spatial concentration of commercial buildings in Chicago. The left panel presents the user query and the agent-generated interpretation, which identifies the Michigan–Wabash area as a dense commercial and mixed-use district based on a 1950 Sanborn map. The right panel shows the corresponding source sheet, with street highlighted in blue, selected commercial building highlighted in orange, and the sheet number indicated in green.

Citation enforcement requires each generated factual answer to include a traceable source identifier, whereas factual grounding requires the substantive claims in the answer to be supported by the cited source evidence. Because citation enforcement alone does not guarantee factual grounding, we conducted five rounds of random sampling, selecting 50 questions in each round: 40 answerable questions and 10 questions requesting attributes not represented in the indexed evidence. We manually audited the resulting 250 questions. For each times, we verified that the cited sheet contained the designated target record and assessed whether every substantive claim was supported by the cited Sanborn evidence. Factual grounding was evaluated using two measures: citation correctness, indicating whether the cited sheet supported the answer, and answer correctness, requiring all response be supported by the cited evidence without unsupported additions or contradictions. All 200 answerable responses cited the correct source sheet, yielding a citation-correctness rate of 100%. The five round-specific answer-correctness rates were 92.1%, 94.6%, 96.8%, 97.5%, and 97.0%, with a mean of 95.6% and a sample standard deviation of 2.25 percentage points. The remaining 50 questions requested attributes are not represented in the indexed evidence, and all 50 questions correctly refrained from providing unsupported answers, corresponding to an abstention-correctness rate of 100.0% (50/50).

6.4.3 Cross-Year Query Evaluation Results

ArchiveLens first separated the indexed records by publication year, creating distinct 1918 and 1920 subsets, and then independently aggregated the corresponding building records within each subset. As shown in Figure 11, the 1918 subset contained 8,791 indexed building records distributed across 122 sheets with eligible building data. ArchiveLens classified 5,695 records as residential (64.8%), 419 as commercial (4.8%), and 157 as industrial (1.8%); 2,520 records (28.7%) were outside the three requested categories. The 1920 subset contained 7,010 building records across 105 sheets, including 3,357 residential records (47.9%), 285 commercial records (4.1%), and 90 industrial records (1.3%). A further 3,278 records (46.8%) were outside the requested categories. These results demonstrate that ArchiveLens can translate a natural-language multi-year query into explicit temporal filters, normalize historical utilization codes, perform year-stratified corpus aggregation, and return a structured comparison without allowing one year or map sheet to dominate the evidence. The reported differences should not be interpreted as longitudinal changes in Chicago's built environment. The 1918 and 1920 map subsets contain different numbers of sheets and may represent different or non-overlapping spatial extents. ArchiveLens therefore labels the output as a cross-sectional comparison of the available indexed records.

**User Query: Cross-Year Comparison**

**User Question:**
Compare the extracted residential, commercial, and industrial records between 1918 and 1920?

**AI Agent Response: Comparisons Analysis across years**

**AI Agent Answer:**
Cross-year building-use comparison

| Year | Indexed sheets | Building records | Residential | Commercial | Industrial | Outside requested categories |
|---|---|---|---|---|---|---|
| 1918 | 122 | 8,791 | 5,695 (64.8%) | 419 (4.8%) | 157 (1.8%) | 2,520 (28.7%) |
| 1920 | 105 | 7,010 | 3,357 (47.9%) | 285 (4.1%) | 90 (1.3%) | 3,278 (46.8%) |

**Figure 11.** Evaluation of ArchiveLens using a cross-year building-utilization query. The agent separates records from 1918 and 1920, aggregates residential, commercial, and industrial utilization categories, and reports counts and percentages.

6.4.4 Retrieval Strategy Comparison Results

ST-RAG achieved the highest overall reference-candidate retrieval performance, retrieving at least one preidentified supporting record for 47 of the 50 questions (94%). Keyword/symbolic filter-only retrieval recovered reference candidates for 42 questions (84%), whereas dense-only retrieval did so for 35 questions (70%). ST-RAG therefore exceeded dense-only retrieval by 24% and filter-only retrieval by 10% Table 6.

Table 6. Reference-candidate retrieval and metadata compliance across the 50-question benchmark set.

| Retrieval configuration | Questions with a reference candidate | Reference-candidate retrieval rate | Metadata-inconsistent records |
|---|---|---|---|
| Dense-only | 35/50 | 70% | 119/250 (47.6%) |
| Keyword/symbolic filter-only | 42/50 | 84% | 0/250 (0%) |
| ST-RAG | 47/50 | 94% | 0/250 (0%) |

The three configurations also differed in how consistently they followed the metadata requirements specified in each query. Dense-only retrieval returned 119 metadata-inconsistent records, representing 47.6% of its 250 retrieved records. In contrast, neither keyword/symbolic filter-only retrieval nor ST-RAG returned records that violated the parsed city, year, volume, or sheet constraints.

Reference-candidate successful rate also varied across query groups (Table 7). ST-RAG recovered supporting candidates for all ten industrial-function queries, storage-and-commerce queries, and localization queries, compared with nine in each category for filter-only retrieval. For map-specific localization controls, filter-only and ST-RAG perform better than dense-only, whereas dense-only retrieval only successfully retrieves five records.

Table 7. Number of questions with at least one retrieved reference candidate by query group.

| Question group | Questions | Dense-only | Keyword/symbolic filter-only | ST-RAG |
|---|---|---|---|---|
| Semantic extraction | 10 | 6 | 6 | 8 |
| Public facilities | 10 | 7 | 9 | 9 |
| Industrial functions | 10 | 8 | 9 | 10 |
| Storage and commerce | 10 | 9 | 9 | 10 |
| Map-specific localization controls | 10 | 5 | 9 | 10 |
| Total | 50 | 35 | 42 | 47 |

Inspection of individual responses revealed differences in evidence selection and answer completeness that were not fully captured by the aggregate retrieval metric. A question that requested a building associated with metal refining and an underground gasoline tank in the 1918 Chicago sheets: dense-only and filter-only retrieval failed to retrieve the record, but ST-RAG successfully retrieved the *Metal Block Corporation* record and returned its name, material, and supporting sheet. Table 8 and Figure 12 illustrate that the dense-only retrieval configuration returned FORT DEARBORN METAL CO. METAL REFINERY from a 1920 sheet. This error reflects the core characteristic of dense retrieval: it prioritizes semantic similarity between the query and indexed records, successfully identifying a conceptually related building associated with *metal refining*, but it does not explicitly enforce metadata constraints such as publication year. As a result, the retrieved evidence was semantically plausible but temporally inconsistent with the query.

Table 8. Response based on Dense-only filter

| | |
|---|---|
| Question | In the 1918 Chicago sheets, identify an industrial building associated with metal refining and an underground gasoline tank. Give its original name, material, and sheet source. |
| Dense-only | FORT DEARBORN METAL CO. METAL REFINERY; Brick (Pink)<br>chicago_ill_1920_vol_23_p15__NjYLmH_00 |

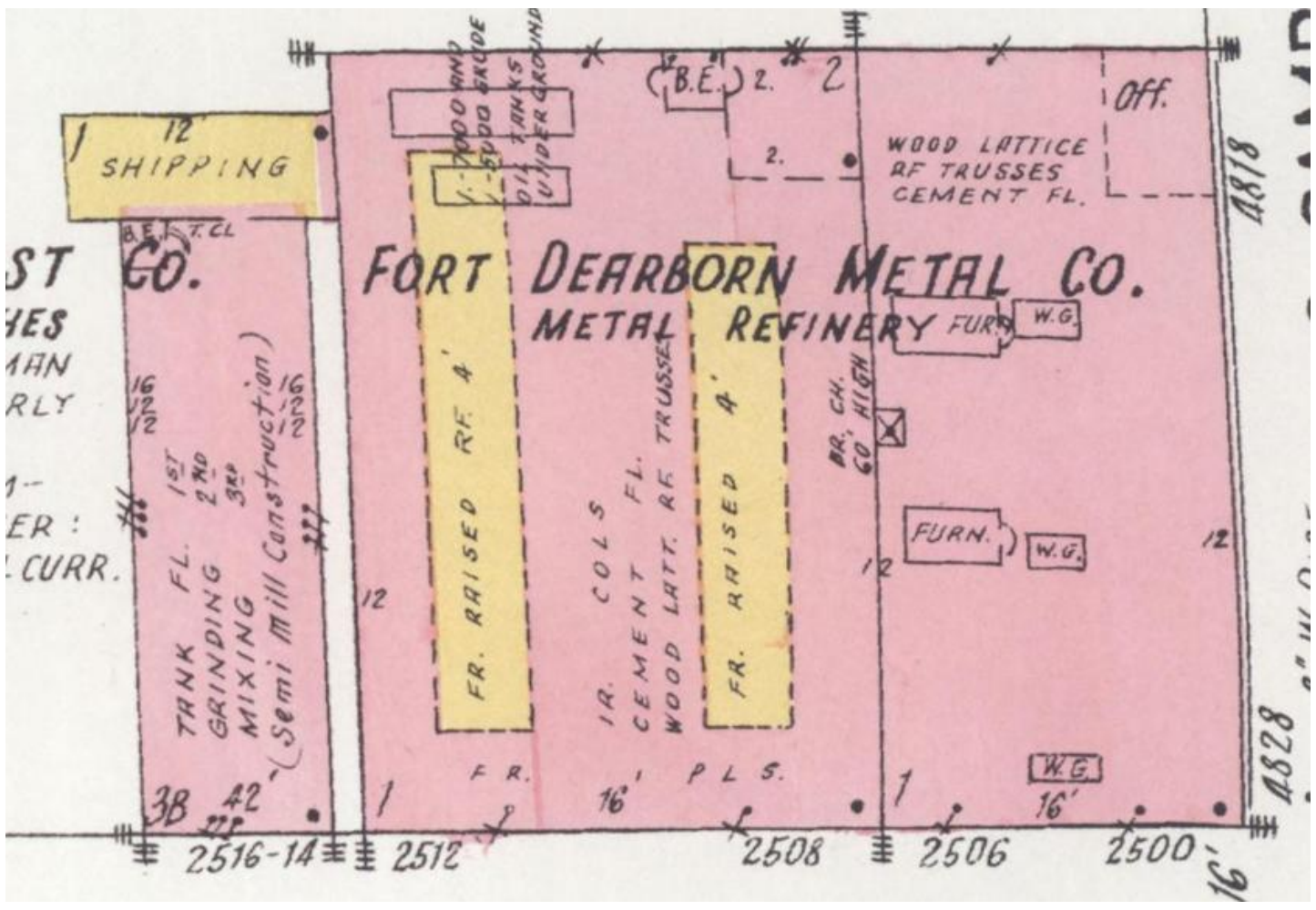

Figure 12. Dense-only retrieval of an industrial building associated with metal refining from an incorrect year.

The keyword/symbolic filter-only retrieval configuration returned *J.A. Ross & Co. Concrete Block Works* from the 1918 sheet (Table 9). This result satisfied the temporal constraint and included an underground gasoline tank note, but it did not correspond to a building specifically associated with *metal refining* (Figure 13). This behavior reflects the limitation of symbolic filtering: the method effectively enforces explicit metadata and phrase constraints, but without semantic similarity it may retrieve records that match isolated query terms or notes while failing to capture the broader functional meaning of the query. In this case, the presence of an underground gasoline tank note and industrial context was sufficient to elevate a record that was lexically compatible but semantically mismatched.

Table 9. Response based on filter-only retrieval

| | |
|---|---|
| Question | In the 1918 Chicago sheets, identify an industrial building associated with metal refining and an underground gasoline tank. Give its original name, material, and sheet source. |
| Filter-only | JA ROSS & CO. BLOCK WORKS；Frame (Yellow)<br>chicago_ill_1918_vol_21_p78__y60fu8_00 |

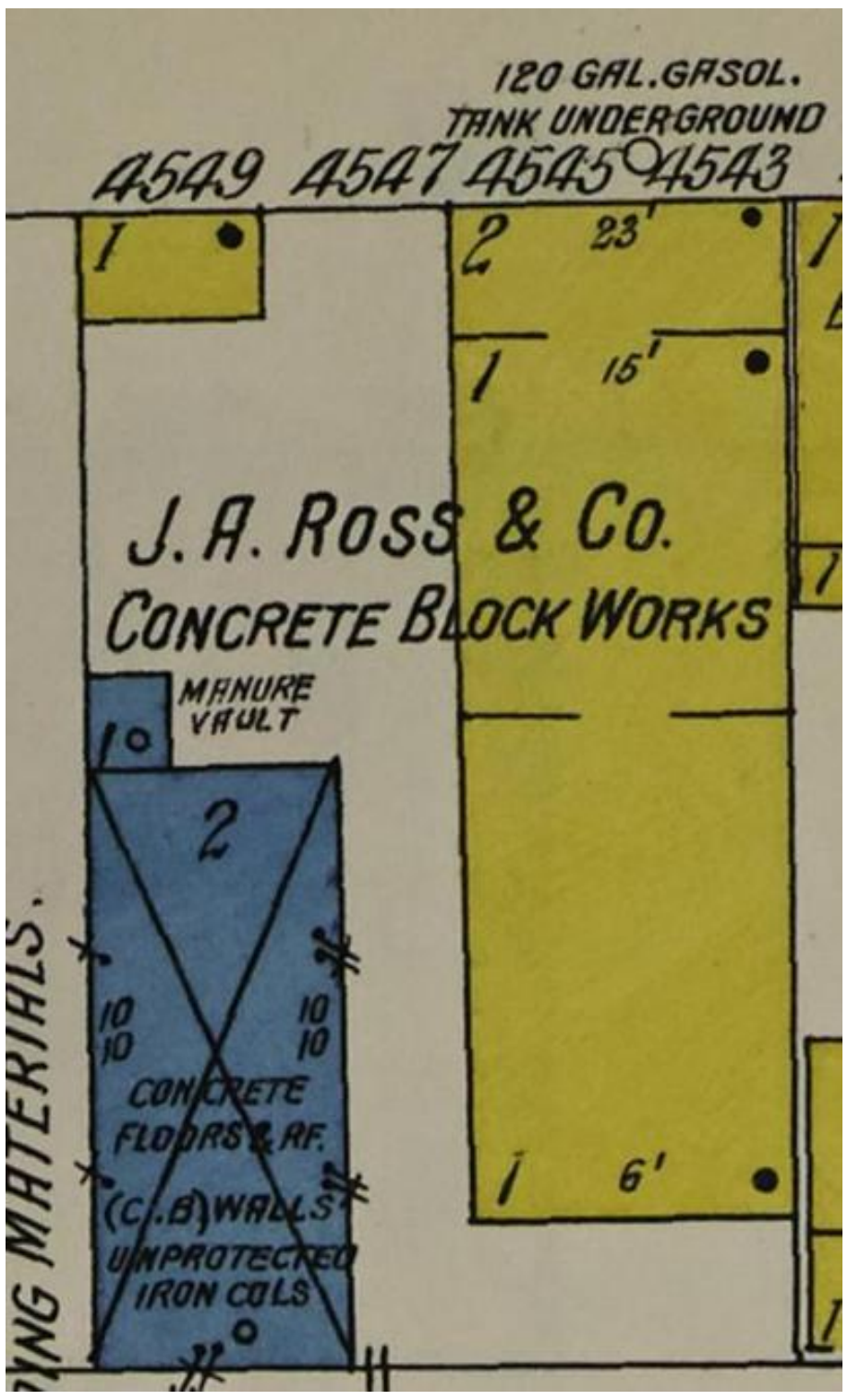

Figure 13. Filter-only retrieval of an industrial building associated from correct year but unrelated to metal refining.

ST-RAG, in contrast, returned *Metal Block Corporation*, a 1918 building record whose evidence explicitly supported all requested elements: the industrial function (*metal refining*), the building material (*brick*), and the note indicating an underground gasoline tank. This example demonstrates the advantage of hybrid retrieval. By combining semantic relevance with explicit temporal and metadata filtering, ST-RAG was able to avoid the temporally inconsistent error of dense-only retrieval and the semantically incomplete match produced by filter-only retrieval. The case therefore illustrates that neither semantic similarity nor symbolic filtering alone was sufficient for this query; accurate retrieval required the integration of both.

Table 10. Response based on ST-RAG retrieval

| | |
|---|---|
| Question | In the 1918 Chicago sheets, identify an industrial building associated with metal refining and an underground gasoline tank. Give its original name, material, and sheet source. |
| ST-RAG | Metal Block Corporation; Material: Brick (Pink); METAL REFINING, WOOD ROOF TRUSS, CONCRETE FLOOR, 450 GAL. GASOL. TANK UNDERGR.<br>chicago_ill_1918_vol_21_p114__e9dzwZ_00 |

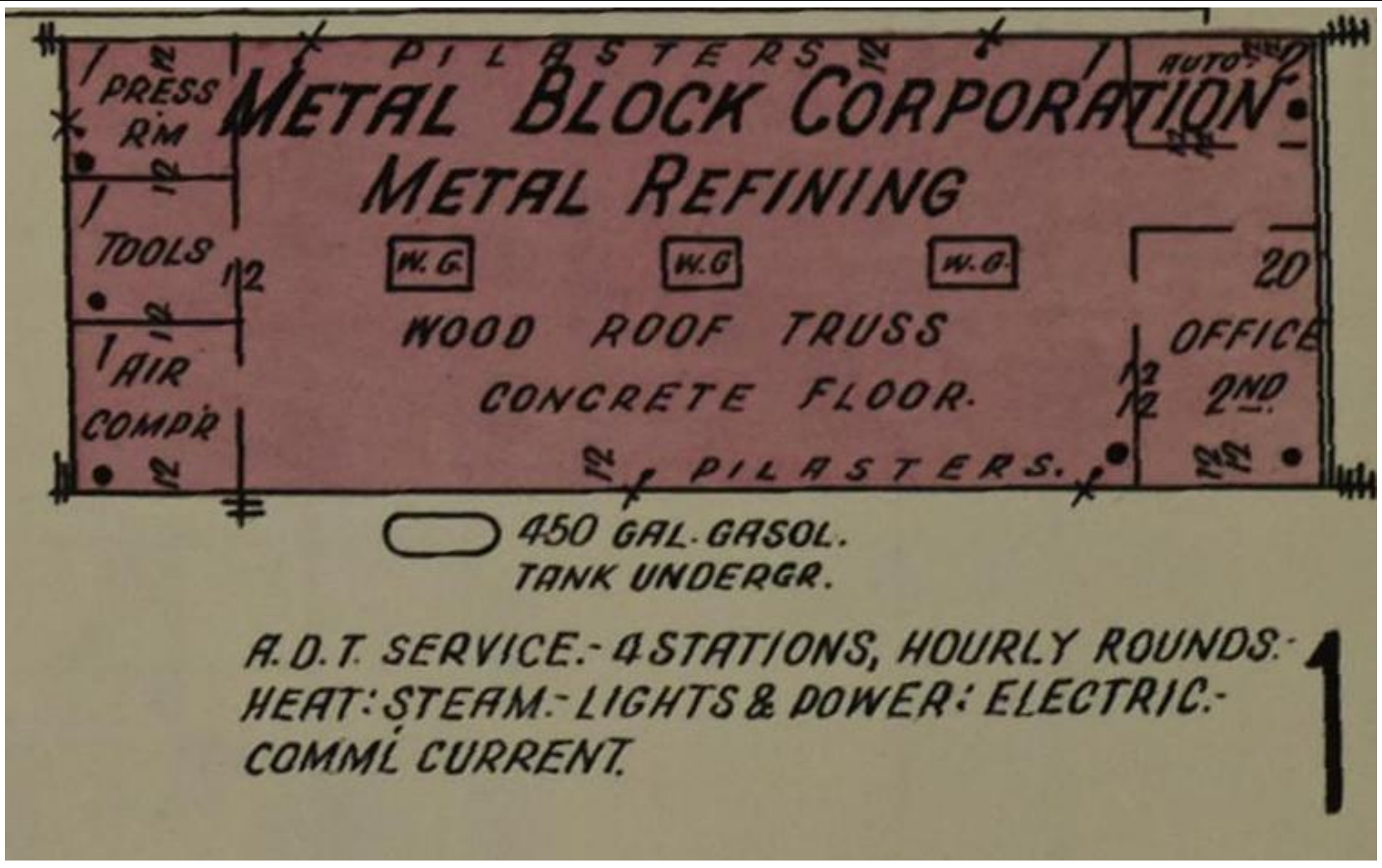

Figure 14. ST-RAG retrieval of an industrial building from the correct year and associated with metal refining.

## 7. Discussion

### 7.1 Transforming Historical Cartography into Entity-Centric Geospatial Records

Historical geographic data, including cartographic, textual, and photographic sources, are inherently unstructured, heterogeneous, and fragmented. The primary contribution of this research extends beyond the digitization of Sanborn Fire Insurance Maps by demonstrating how complex historical archives can be systematically transformed into structured, entity-centric geospatial records. ArchiveLens operationalizes this transformation by extracting individual building footprints and linking them with attributes such as construction materials and building uses. In this way, visually encoded map content is converted into discrete, searchable records that can support building-level and large-scale historical spatial analysis. These structured records can also provide a foundation for future historical knowledge graphs and more advanced analyses of urban form and change over time.

### 7.2 Hybrid ST-RAG retrieval

The ST-RAG design contributes a retrieval strategy tailored to historical map archives by combining the complementary strengths of semantic and symbolic retrieval. Dense-only retrieval is effective at identifying conceptually similar records, but the comparison showed that semantic similarity alone can overlook explicit temporal or map-specific constraints, resulting in evidence that is relevant in meaning but inconsistent with the requested year, volume, or sheet. Keyword/symbolic filtering provides stronger control over these constraints, but its reliance on explicit terms can favor lexically matched records that do not fully capture the functional meaning of a query. ST-RAG addresses these limitations by integrating semantic similarity with spatial, temporal, and keyword-based constraints within a single retrieval process. In the controlled comparison, ST-RAG achieved the highest reference-candidate retrieval rate (94%), compared with 84% for filter-only retrieval and 70% for dense-only retrieval, while also returning no metadata-inconsistent records. The contribution of ST-RAG therefore lies not simply in improving retrieval performance, but in balancing conceptual relevance with explicit archival constraints, which is particularly important for historical map collections where records are fragmented across years, volumes, and sheets and where semantically similar evidence may not correspond to the correct historical context

### 7.3 Archival Rigor and the Mitigation of Generative Hallucination

The operationalization of historical data is measured by its accessibility and evidentiary reliability. The final phase of the framework demonstrates how structured historical geospatial records can be seamlessly interrogated using natural language via an AI agent. However, the application of generative AI to historical archives introduces the severe risk of hallucination, which is a critical barrier for disciplines that demand an unbroken chain of evidence. We mitigate this risk at the synthesis layer through the enforcement of a strict "Mandatory Citation" mechanism. By requiring the Large Language Model to append a precise source sheet identifier to any factual claim or structural description, the system is constrained to act strictly as a grounded archival assistant. Visual cross-referencing shows that the semantic details and spatial relationships generated by the agent can be traced back to corresponding evidence in the original cartographic source. This mechanism ensures that the convenience of natural language querying does not compromise the uninterrupted provenance trail required for rigorous academic research.

## 8. Limitations and Future Directions

While the framework demonstrates robust extraction capabilities, its semantic performance is subject to the visual constraints of the source material. Evaluations indicate that Vision-Language Models are highly effective at transcribing typography within larger architectural footprints, but misclassifications increase in densely packed residential outbuildings where cartographers truncated abbreviations or where ink overlapped geometric boundaries.

The deterministic HSV material classifier has important boundary conditions. Its performance in this study reflects the standardized Sanborn color conventions and the visual characteristics of the digitized Chicago corpus; the reported thresholds should not be assumed to transfer unchanged across cities, publication periods, volumes, or archival providers. Although Sanborn maps generally use consistent colors to distinguish construction materials, observed pixel values can vary because of paper aging, pigment fading, staining, scanning illumination, compression, color-profile conversion, handwritten revisions, and grayscale or otherwise non-standard reproduction. These variations may shift a nominal material color outside the predefined HSV ranges or increase overlap between classes. Consequently, applying ArchiveLens to a new collection should include an initial color-fidelity assessment and, where necessary, recalibration using representative reference samples from that collection. This study did not conduct a cross-city or cross-provider sensitivity analysis; therefore, the reported material-classification performance establishes robustness only within the evaluated Chicago corpus and acquisition conditions. Evaluating threshold stability under controlled color perturbations and across independently digitized Sanborn collections remains an important direction for future work.

A limitation of this study is the inconsistent spatial coverage of the Sanborn map corpus across publication years. Although the 445 sheets span 1918, 1920, 1923, and 1950, different years often cover different or non-overlapping parts of Chicago. Therefore, aggregate differences in building counts, materials, or utilization codes should be interpreted as cross-sectional differences in the available records rather than direct longitudinal change. ArchiveLens preserves publication year, source sheet, volume, and spatial context to make these coverage differences explicit, but it does not reconstruct missing historical coverage. Future work should focus on fixed neighborhoods or corridors with overlapping map editions and on cross-year building matching to support more rigorous longitudinal analysis.

A future extension of ArchiveLens could organize the extracted records as a provenance-aware spatio-temporal knowledge graph. Nodes would represent building records, physical buildings, Sanborn sheets, streets, blocks or parcels, construction materials, utilization types, and publication years. Edges would associate building records with their source sheets, materials, utilization codes, and publication years; spatial edges would represent adjacency between buildings, adjacency to streets, and containment within blocks or parcels. Cross-year entity-resolution edges could connect records inferred to represent the same physical building, supporting temporal relations such as persistence, appearance, disappearance, material change, and utilization change. To preserve archival traceability, each node and relation would retain the supporting sheet identifier, while spatially or temporally inferred relations would additionally record their derivation method and confidence. Such a graph could support multi-hop queries across space, time, and building attributes; however, developing and validating cross-year entity matching and temporal relations remain subjects for future work.

## 9. Conclusion

The ArchiveLens framework provides a scalable, three-phase methodology that successfully bridges the gap between raw cartographic imagery and interactive spatial analysis. By coupling deep learning instance segmentation (Phase 1) with localized semantic parsing (Phase 2), the pipeline delineates the physical boundaries of the urban fabric while extracting vital contextual typography. In Phase 3, the integration of these attributes into an intent-aware, hybrid ST-RAG system transforms the data into an interactive, provenance-grounded spatial query engine.

Most importantly, the paradigm established here extends far beyond Sanborn maps. It offers a comprehensive blueprint for the digital humanities and GIScience communities to process inherently fragmented, unstructured historical records into highly structured, traceable formats. By organizing archival data as an entity-centric geospatial database accessible via natural language, this framework empowers researchers to conduct verifiable, large-scale analyses of urban morphology without severing the vital link to the original historical source.

## Appendices

### Appendix A.1 Prompt templates and inference configuration

You are analyzing a full sheet of a Sanborn Fire Insurance Map.
The file name of this map is "{filename}", which often contains highly
accurate hints about the city, state, year, volume, and sheet number
(e.g., "chicago_ill_1950_vol_11_p28" implies Chicago, Illinois, 1950,
Volume 11, Sheet/Page 28).

Combine the textual clues from the filename with the visual information
in the image to extract the following information strictly as a JSON
object:

1) metadata: Extract the following fields:
- "city": City name (infer from filename if blurry).
- "state": State abbreviation or name.
- "year": Map year.
- "volume": Map volume number.
- "sheet": The sheet/page number.
- "north_angle_deg": Approximate north arrow angle if visible visually.

2) road_names: List all street/road names visible on the map.
For each: "name", "confidence" (0.0-1.0), "orientation". Max 40.

3) extra: Provide a concise visual description of the entire map sheet
context (e.g., "Industrial district with large factories",
"Residential area with dense dwellings").

Return ONLY this JSON:

```
{
  "metadata": {
    "city": "",
    "state": "",
    "year": "",
    "volume": "",
```

```
        "sheet": "",
        "north_angle_deg": ""
    },
    "road_names": [
        {
            "name": "",
            "confidence": 0.0,
            "orientation": ""
        }
    ],
    "extra": ""
}
```

Appendix A.2 Building-level attribute extraction

You are given a crop from a Sanborn Fire Insurance Map showing ONE building.
Extract building information using standard Sanborn abbreviations.

Return ONLY this JSON:

```
{
    "utilization": [],
    "utilization_notes": "",
    "name": "",
    "num_storeys": "0",
    "extra": ""
}
```

Rules:

- utilization: List abbreviations
  (D/Dwg=dwelling, F=flat, S=store, A/Auto=garage,
  Po=post office, Sal=saloon, B=basement,
  W=warehouse, Mfg=manufacturing).
- num_storeys: Extract the storey number exactly as written
  (e.g., "1", "2", "2 1/2", "3"). Use "0" if missing.
- name: Building or business name if labeled.
- extra: Transcribe any other useful text found inside the building
  (e.g., "Tin Shop", "Hardw.", "Flats abv").

Appendix A.3 Evidence-grounded synthesis with mandatory citation

You are an expert GIS and urban history assistant.
Answer based ONLY on the supplied ArchiveLens records.

STRICT RULES:

1. Match the exact target building or quoted record phrase in the question. Never substitute a similar building.

2. State only fields explicitly present in that matched record.
   Do not decode abbreviations, infer ownership, infer architects,
   or infer missing attributes.
3. If the exact target or requested field is absent, answer:
   "The requested information is not recorded in the retrieved evidence."
4. Keep the answer concise and directly answer the question.
5. Every factual statement must cite its supporting sheet with the
   exact tag [sheet_stem].
6. A retrieved source is not evidence unless its text explicitly
   supports the statement.

Appendix A.4 Computational environment and run protocol

All VLM extraction and synthesis experiments were executed locally in a CUDA-enabled container on an NVIDIA GeForce RTX 4090 GPU with 24 GB of VRAM. The runtime used PyTorch and Hugging Face Transformers for VLM inference, SentenceTransformers for record encoding, and FAISS for vector indexing and search. Each sheet image and building crop was processed once. Each question in the synthesis benchmark likewise received one deterministic generation. No repeated trials, majority voting, self-consistency decoding, or post hoc selection of generations was performed.